\documentclass{article}
\pdfoutput=1
\usepackage[numbers]{natbib}
\usepackage[preprint]{neurips_2020}
\usepackage[utf8]{inputenc} % allow utf-8 input
\usepackage[T1]{fontenc}    % use 8-bit T1 fonts
\usepackage{hyperref}       % hyperlinks
\usepackage{url}            % simple URL typesetting
\usepackage{graphicx}
\usepackage[ruled,linesnumbered,vlined]{algorithm2e}
\usepackage{booktabs}       % professional-quality tables
\usepackage{amsmath,amsfonts,amssymb}
\usepackage{nicefrac}       % compact symbols for 1/2, etc.
\usepackage{microtype}      % microtypography
\usepackage{multirow}
\usepackage{amsmath}
\usepackage{float}
\usepackage{listings}
\usepackage{markdown}
\usepackage{subfig}
\usepackage{tabularx}
\usepackage[Q=yes]{examplep}
\usepackage[table,xcdraw]{xcolor}
\usepackage{hyperref}
\usepackage{algpseudocode}
\usepackage{xspace}
\usepackage{xcolor}
\usepackage{footnote}
\makesavenoteenv{tabular}
\makesavenoteenv{table}

\newcommand{\codename}{GShard\xspace}

\newcommand{\MoE}[2]{MoE(#1E,~#2L)}
 
\newcommand{\TPUv}{TPU v3\xspace}
\usepackage{etoolbox}
\gappto{\UrlBreaks}{\UrlOrds}

\title{GShard: Scaling Giant Models with Conditional Computation and Automatic Sharding}

\author{%
Dmitry Lepikhin\\\texttt{lepikhin@google.com}\And
HyoukJoong Lee\\\texttt{hyouklee@google.com}\And
Yuanzhong Xu\\\texttt{yuanzx@google.com}\And
Dehao Chen\\\texttt{dehao@google.com}\And
Orhan Firat\\\texttt{orhanf@google.com}\And
Yanping Huang\\\texttt{huangyp@google.com}\And
Maxim Krikun\\\texttt{krikun@google.com}\And
Noam Shazeer\\\texttt{noam@google.com}\And
Zhifeng Chen\\\texttt{zhifengc@google.com} \\~
}

\begin{document}

\maketitle

% \section{Overall}
% \begin{itemize}
% \item New Green Giant model
% \begin{enumerate}
% \item 600B (largest ever): model arch (MoE+Xformer); conditional computation;
% \item MT SOTA BLEU result.
% \item Trains fast: sample efficiency / cost efficiency.
% \end{enumerate}

% \item Infra enabler
% \begin{enumerate}
% \item XLA sharding: compiler side does the heavy-lifting for scalable codegen/parallelization/scaling.
% \item We leveraged TPU's ICI hardware. (some experiments/measurement to showcase).
% \item SPMD vs. MPMD
% \item all-to-all efficiency.
% \end{enumerate}
% \end{itemize}

% MoE with SPMD Partitoner:
%     - 1 Introduction
%     	and related scaling work
%     	- Batch-, Model- and Spatial- partitioning
%     	- Dense scaling
%     	- Sparse scaling
%     - 2 Sparse scaling with Mixture of Experts
%     - 3 Experiments
%         - Multilingual Machine Translation
%         	- Model Architecture
%         	- Dataset
%         	- Results
%         		- Quality
%         		- Efficiency
%     - 4 XLA TPU Model Parallelism Infrastructure
%     	http://doc/1T6KlfqTUp5riV1Cq-IovB8Cnoi8rMII_AmOgUgHol6E#heading=h.sr6zac68v8m6
%     	- SPMD Partitioning
%     		http://go/spmd-partitioning
%     - 5 Conclusion
%     Appendix:
%     	- MoE notation and meta-code
%       - SPMD Partitioning details 
%           - Showcase lightweight annotation syntax
%           - Showcase spatial partitioning with large example and cnn models implemented with go/spmd-partitioning.
%             We made an effort to list different partitioning in the Intro section.    

\begin{abstract}
Neural network scaling has been critical for improving the model quality in many real-world machine learning applications with vast amounts of training data and compute. Although this trend of scaling is affirmed to be a sure-fire approach for better model quality, there are challenges on the path such as the computation cost, ease of programming, and efficient implementation on parallel devices. \codename is a module composed of a set of lightweight annotation APIs and an extension to the XLA compiler. It provides an elegant way to express a wide range of parallel computation patterns with minimal changes to the existing model code. GShard enabled us to scale up multilingual neural machine translation Transformer model with Sparsely-Gated Mixture-of-Experts beyond 600 billion parameters using automatic sharding. We demonstrate that such a giant model can efficienctly be trained on 2048 TPU v3 accelerators in 4 days to achieve far superior quality for translation from 100 languages to English compared to the prior art.

\end{abstract}
% Types:
% 	- Data-Parallelism (batch-splitting)
% 		- Universal, agnostic to model architecture
% 		- No model-splitting (no large models)
% 		- No example-splitting (no large examples)
% 	- Model-Parallelism (model-splitting)
% 		- Efficiency generally depends on model architecture
% 		- "Efficient and task-independent model parallelism, we introduce GPipe" (cite gpipe)
% 			- Sequence of layers is split between cores, and pipelined
% 			- Bubble
% 	- Spatial Splitting of large inputs (images/video)
% 		- Enables giant examples

\section{Introduction}
\label{sec:intro}

% Motivation
%Theoretical and empirical results continually show that neural network scaling is highly correlated with machine learning model quality
Scaling neural networks brings dramatic quality gains over a wide array of machine learning problems ~\cite{arora2018optimization,frankle2018lottery,kaplan2020scaling, devlin2018bert,mahajan2018exploring,gpt32020}. 
%given sufficiently large training sets and compute. 
For computer vision, increasing the model capacity has led to better image classification and detection accuracy for various computer vision architectures \cite{he2016deep, he2016identity, ghiasi2019fpn}. Similarly in natural language processing, scaling Transformers \cite{vaswani2017attention} yielded consistent gains on language understanding tasks \cite{devlin2018bert,raffel2019exploring,brown2020language}, cross-lingual down-stream transfer \cite{devlin2018bert,conneau2019unsupervised} and (massively-)multilingual neural machine translation~\cite{arivazhagan2019massively, gpipe19, shazeer2017outrageously}. 
%And similarly in reinforcement learning with self-play
%experiments with multilingual machine translation demonstrated that models with larger capacity not only scored better translation quality, but also transferred better among different language pairs when multitasking.
This general tendency motivated recent studies to scrutinize the factors playing a critical role in the success of scaling \cite{advani2017highdimensional,hestness2017deep,Hestness_2019,Geiger_2020,kaplan2020scaling}, including the amounts of training data, the model size, and the computation being utilized as found by past studies. While the final model quality was found to have a power-law relationship with the amount of data, compute and model size \cite{hestness2017deep,kaplan2020scaling}, the significant quality gains brought by larger models also come with various practical challenges. \textit{Training efficiency} among the most important ones, which we define as the amount of compute and training time being used to achieve a superior model quality against the best system existed, is oftentimes left out. 

% figure-1, the figure we want people to retweet about.	It should only cover the relationship among model quality, capacity and training time, without details about network architecture, compute topology, since most of the model/experiments details are covered in section 2.
\begin{figure}[ht!]
\centering
\includegraphics[width=0.9\textwidth]{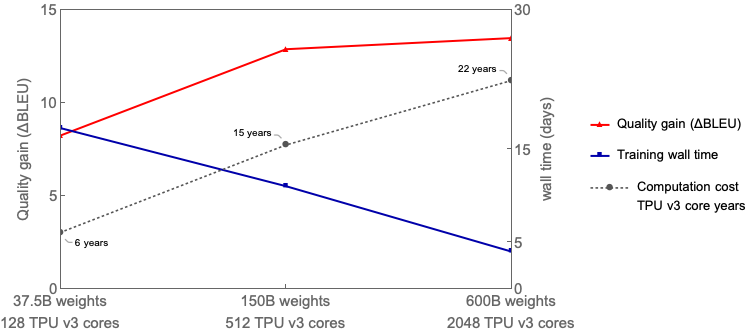}
\caption{Multilingual translation quality (average $\Delta$BLEU comparing to bilingual baselines) improved as MoE model size grows up to 600B, while the end-to-end training cost (in terms of \TPUv core-year) only increased sublinearly. Increasing the model size from 37.5B to 600B (16x), results in computation cost increase from 6 to 22 years (3.6x). The 600B parameters model that achieved the best translation quality was trained with 2048 \TPUv cores for 4 days, a total cost of 22 \TPUv core-years. In contrast, training all 100 bilingual baseline models would have required 29 \TPUv core-years. Our best quality dense single Transformer model (2.3B parameters) achieving $\Delta$BLEU of 6.1, was trained with GPipe~\cite{gpipe19} on 2048 \TPUv cores for 6 weeks or total of 235.5 \TPUv core-years.}
\label{fig:intro_quality_vs_time}
\end{figure}

% Punchline --> moting this to related work of machine translation.
%\todo{lepikhin to simplify the Punchline: set up what we want to do, at the moment paragraph is not readable}
%In this study, we strive for improving the model quality while together with being training efficient. We achieve progressively better quality models by scaling neural sequence models beyond 600 billion parameters, and realize training efficiency by algorithmic improvements in model design and systems design. For model design, we make extensive use of conditional computation~\cite{bengio2015conditional,shazeer2017outrageously,Elbayad2020DepthAdaptiveT,bapna2020controlling}, which premises that the examples should be routed within the network by activating an input dependent subnetwork. The routing depends (or conditions) on certain criterion and without the loss of generality, can be any of the following: estimated difficulty of the example \cite{lugosch2020surprisaltriggered}, available computation budget \cite{Elbayad2020DepthAdaptiveT,bapna2020controlling}, or more generally a learned criterion with sparsity induced mixture of experts \cite{shazeer2017outrageously}. We extend sparsely gated mixture of experts \cite{shazeer2017outrageously} due to its flexibility and ease of scaling to state of the art neural sequence models, Transformers \cite{vaswani2017attention}, to satisfy training efficiency. For system design ...\todo{can someone from section 3 or 5 continue this paragraph to give a brief and high level rationale for systems design contribution of the paper please?}.

% Complexity.
\subsection{Practical Challenges for Scaling}
Here we enumerate major practical challenges faced especially when training massive-scale models that are orders of magnitude larger than the capacity limit of a single accelerator memory (e.g., GPUs or TPUs). 
\paragraph{Architecture-specific model parallelism support} There is a lack of support for efficient model parallelism algorithms under commonly used deep learning frameworks such as TensorFlow~\cite{abadi2016tensorflow} and PyTorch~\cite{pytorch2017}. Naive model parallelism with graph partition is supported but it would lead to severe under-utilization due to the sequential dependency of the network and gradient based optimization. In order to scale up the existing models efficiently, users typically need to invest a lot of engineering work, for example, migrating the model code to special frameworks~\cite{shazeer2018mesh,gpipe19}.   

\paragraph{Super-linear scaling of computation cost vs model size} Straightforward scaling of the mode size by increasing the depth or width~\cite{gpt32020, gpipe19} generally results in at least linear increase of training step time. Model parallelism by splitting layer weights and computation across multiple devices generally becomes necessary, leading to network communication overhead and device under-utilization. Device under-utilization stems from imbalanced assignment and sequential dependencies of the underlying neural network. This super-linear relationship between the computation cost and the model size can not be resolved by simply using more devices, making training massive models impractical.

\paragraph{Infrastructure scalability for giant model representation} A naive graph representation for the massive-scale model distributed across thousands of devices may become a bottleneck for both deep learning frameworks and their optimizing compilers. For example, adding $D$ times more layers with inter-op partitioning or increasing model dimensions with intra-op partitioning across $D$ devices may result in a graph with $O(D)$ nodes. Communication channels between devices could further increase the graph size by up to $O(D^2)$ (e.g., partitioning gather or transpose). Such increase in the graph size would result in an infeasible amount of graph building and compilation time for massive-scale models.

\paragraph{Non-trivial efforts for implementing partitioning strategies} Partitioning a model to run on many devices efficiently is challenging, as it requires coordinating communications across devices. For graph-level partitioning, sophisticated algorithms~\cite{gpipe19, harlap2018pipedream} are needed to reduce the overhead introduced by the sequential dependencies between different partitions of graphs allocated on different devices. For operator-level parallelism, there are different communication patterns for different partitioned operators, depending on the semantics, e.g., whether it needs to accumulate partial results, or to rearrange data shards. According to our experience, manually handling these issues in the model requires substantial amount of effort, given the fact that the frameworks like TensorFlow have a large sets of operators with ad-hoc semantics. In all cases, implementing model partitioning would particularly be a burden for practitioners, as changing model architecture would require changing the underlying device communications, causing a ripple effect.

%\paragraph{Non-trivial efforts for operator-level partitioning} Partitioning a model with giant tensors to run on many devices is challenging, as it requires adding communications across devices. Each partitioned operator has a different communication pattern depending on its semantics, e.g., whether it needs to accumulate partial results, or to rearrange data shards. Manually handling these issues in the model would require lots of effort, as frameworks like TensorFlow has a large set of operators with ad hoc semantics. Manual model partitioning would particularly be a burden for research experiments, as changing model architecture would require changing the underlying device communications.

\subsection{Design Principles for Efficient Training at Scale} \label{subsec:design} 
In this paper, we demonstrate how to overcome these challenges by building a $600$ billion parameters sequence-to-sequence Transformer model with Sparsely-Gated Mixture-of-Experts layers, which enjoys sub-linear computation cost and $O(1)$ compilation time. We trained this model with $2048$ \TPUv devices for $4$ days on a multilingual machine translation task and achieved far superior translation quality compared to prior art when translating $100$ languages to English with a single non-ensemble model. We conducted experiments with various model sizes and found that the translation quality increases as the model gets bigger, yet the total wall-time to train only increases sub-linearly with respect to the model size, as illustrated in Figure~\ref{fig:intro_quality_vs_time}. 
%We not just train and stop at 4 days. We had a prefix epochs number. 
To build such an extremely large model, we made the following key design choices.

%  we make extensive use of conditional computation~\cite{bengio2015conditional,shazeer2017outrageously,Elbayad2020DepthAdaptiveT,bapna2020controlling}, which premises that the examples should be routed within the network by activating an input dependent subnetwork. The routing depends (or conditions) on certain criterion and without the loss of generality, can be any of the following: estimated difficulty of the example \cite{lugosch2020surprisaltriggered}, available computation budget \cite{Elbayad2020DepthAdaptiveT,bapna2020controlling}, or more generally a learned criterion with sparsity induced mixture of experts \cite{shazeer2017outrageously}. We extend sparsely gated mixture of experts \cite{shazeer2017outrageously} due to its flexibility and ease of scaling to state of the art neural sequence models, Transformers \cite{vaswani2017attention}, to satisfy training efficiency.
\paragraph{Sub-linear Scaling} First, model architecture should be designed to keep the computation and communication requirements sublinear in the model capacity. Conditional computation~\cite{bengio2015conditional,shazeer2017outrageously,Elbayad2020DepthAdaptiveT,bapna2020controlling} enables us to satisfy training and inference efficiency by having a sub-network activated on the per-input basis. Scaling capacity of RNN-based machine translation and language models by adding Position-wise Sparsely Gated Mixture-of-Experts (MoE) layers~\cite{shazeer2017outrageously} allowed to achieve state-of-the-art results with sublinear computation cost. We therefore present our approach to extend Transformer architecture with MoE layers in Section~\ref{sec:model}.

% RNN-based machine translation and language models by adding Position-wise Sparsely Gated Mixture-of-Experts (MoE) layers~\cite{shazeer2017outrageously}, resulted in significant quality improvements with a sub-linear cost.

% \todo{lepikhin: move prior work here, then explain why by describing the property} Sparse activation of a subnetwork on the per-input basis with conditional computation becomes fundamental to this design and allows us to avoid super-linear costs of dense scaling. We formulate the new computation cost with conditional computation as follows:
% $$
% \text{Computation cost} \propto \text{Average size of the subnetwork per input} \times \text{Inputs per batch}
% $$
% For instance, drastically increasing capacity of RNN-based machine translation and language models by adding Position-wise Sparsely Gated Mixture-of-Experts (MoE) layers~\cite{shazeer2017outrageously}, resulted in significant quality improvements with a sub-linear cost. 

\paragraph{The Power of Abstraction}
Second, the model description should be separated from the partitioning implementation and optimization. This separation of concerns let model developers focus on the network architecture and flexibly change the partitioning strategy, while the underlying system applies semantic-preserving transformations and implements efficient parallel execution. To this end we propose a module, \codename, which only requires the user to annotate a few critical tensors in the model with partitioning policies. It consists of a set of simple APIs for annotations, and a compiler extension in XLA~\cite{xla} for automatic parallelization. Model developers write models as if there is a single device with huge memory and computation capacity, and the compiler automatically partitions the computation for the target based on the annotations and their own heuristics. We provide more annotation examples in Section~\ref{sec:spmd_annotation}.

\begin{figure}[t!]
\centering
    \subfloat[MPMD Partition]{
      \includegraphics[width=0.49\textwidth]{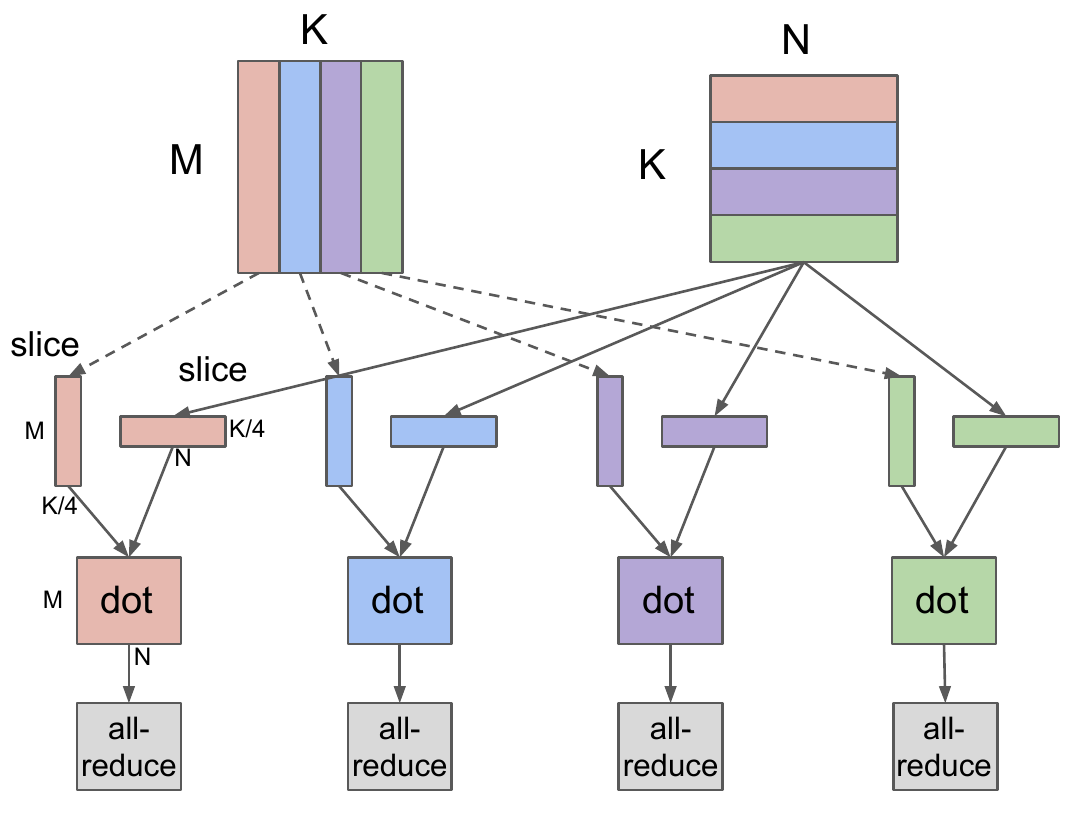}
      \label{fig:spmd_mpmd_comparison:mpmd}
    }
    \qquad
    \subfloat[SPMD Partition]{
      \includegraphics[width=0.40\textwidth]{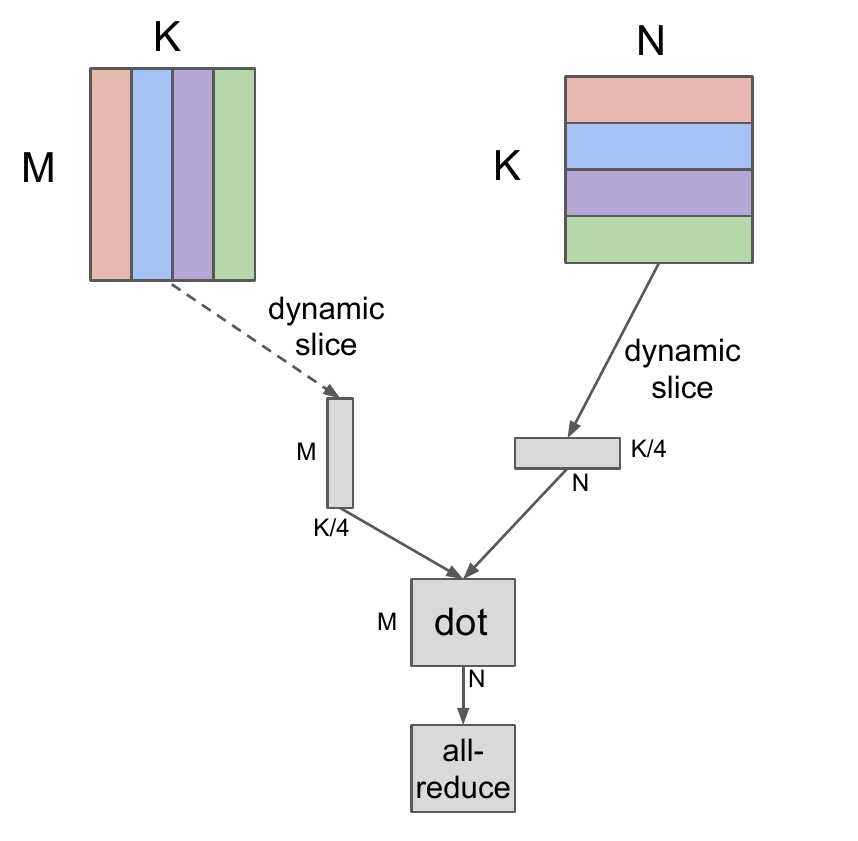}
      \label{fig:spmd_mpmd_comparison:spmd}
    }
\caption{Comparison between MPMD and our proposed SPMD partitioning of a Dot operator ($[M,K] \times [K,N] = [M,N]$) across 4 devices. In this example, both operands are partitioned along the contracting dimension $K$, where each device computes the local result and globally combines with an AllReduce. MPMD partitioning generates separate operators for each device, limiting its scalability, whereas SPMD partitioning generates one program to run on all devices. Note that the compilation time with our SPMD partitioning is not-dependent of the number of devices being used.}
\label{fig:spmd_mpmd_comparison}
\end{figure}

\paragraph{Scalable Compilers} Third, the system infrastructure, including the computation representation and compilation, must scale with thousands of devices for parallel execution. For example, Figure~\ref{fig:spmd_mpmd_comparison} illustrates two different ways of partitioning a dot-product operation across 4 devices (color-coded). Notice that with the usual MPMD (Multiple Program Multiple Data) approach in Figure~\ref{fig:spmd_mpmd_comparison:mpmd} scaling becomes more challenging since the number of nodes in the graph  increases linearly with the number of devices. Instead, we developed a compiler technique for SPMD (Single Program Multiple Data) transformation that generates a single program to run on all devices, keeping the compilation time constant independent of the number of devices, as illustrated in Figure~\ref{fig:spmd_mpmd_comparison:spmd}.  We will discuss our SPMD framework in more details in Section~\ref{sec:partitioner}.

The rest of the paper is organized as the following. Section~\ref{sec:model} describes our Transformer architecture with Sparsely-Gated MoE layer in more details. Section~\ref{sec:framwork}  introduces our development module \codename. 
Section~\ref{sec:experiments} demonstrates the application of our mixture of expert models on the multilingual machine translation task over $100$ language pairs. Section~\ref{sec:performance} has performance and memory measurements of our implementation. Section~\ref{sec:related} discusses related work. 
% Relevance

\section{Model}
\label{sec:model}

\newcommand{\FLOPS}{\text{Compute}}  % consider "number of tokens per batch"
\newcommand{\batchsize}{\text{batch\,size}}  % consider "number of tokens per batch"
\newcommand{\numberofweights}{}

\newcommand{\Expert}{\text{FFN}}
\newcommand{\Gate}{\text{GATE}}
\newcommand{\WG}{wg}
\newcommand{\WI}{wi}
\newcommand{\WO}{wo}

\subsection{Sparse scaling of the Transformer architecture}
\label{sec:sparse}
% MoE motivation is covered in Sublinear design principle with conditional computation.
%
% Please sync w/Dima before updating.

% Was covered in challenges in the into AKA "super-linear" scaling
% Increasing the number of layers and the model dimension of the Transformer model have been shown as an effective way to improve the multilingual machine translation quality~\cite{gpipe19}. However, this \textit{dense} model scaling results in a computation cost proportional to the model size, and becomes impractical beyond few tens of billions of weights.

\begin{figure}[t!]
\begin{center}
\includegraphics[width=0.95\textwidth]{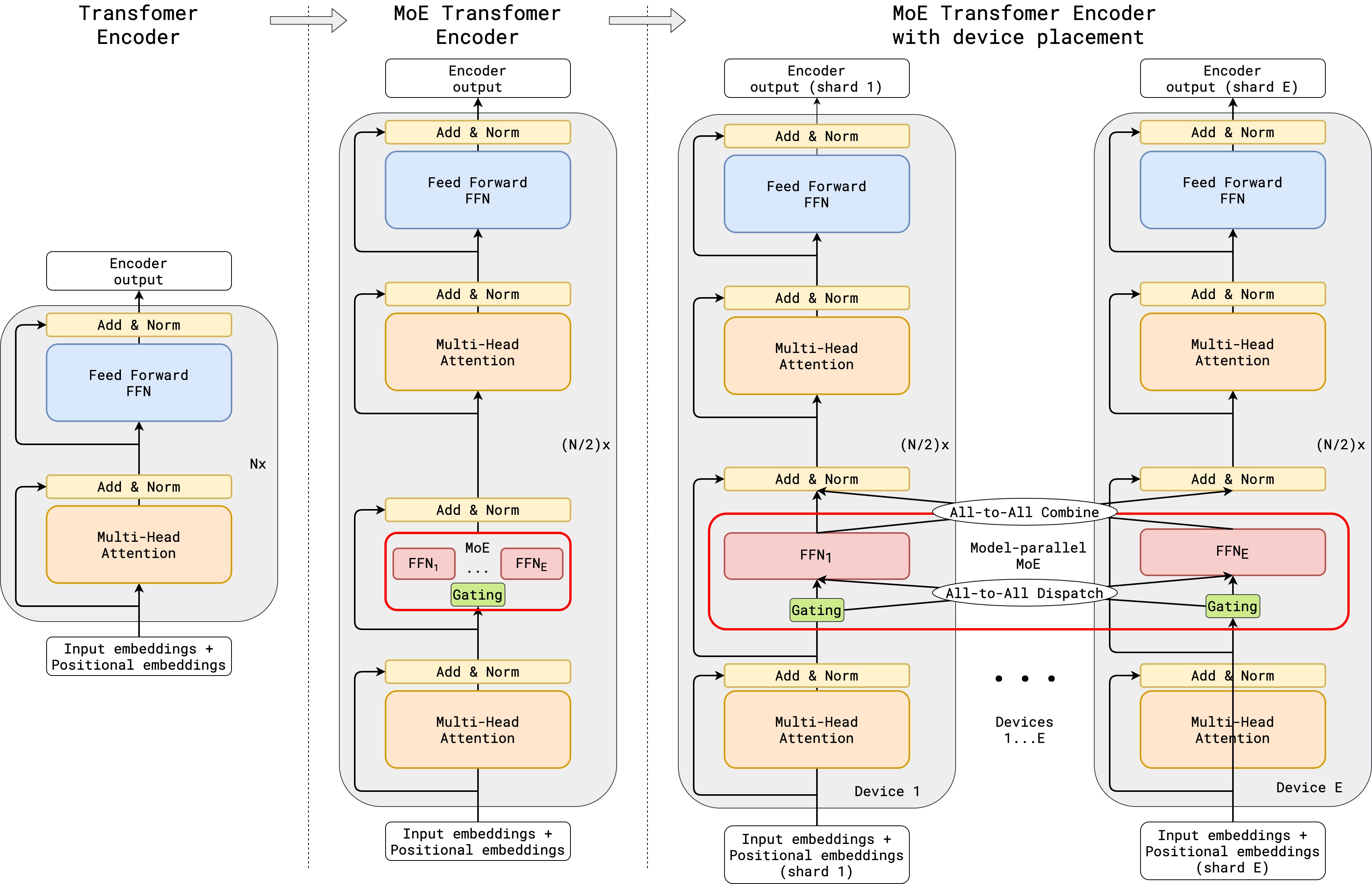}
\caption{Illustration of scaling of Transformer Encoder with MoE Layers. The MoE layer replaces the every other Transformer feed-forward layer. Decoder modification is similar. (a) The encoder of a standard Transformer model is a stack of self-attention and feed forward layers interleaved with residual connections and layer normalization. (b) By replacing every other feed forward layer with a MoE layer, we get the model structure of the MoE Transformer Encoder. (c) When scaling to multiple devices, the MoE layer is sharded
across devices, while all other layers are replicated.
}
\label{fig:moe_architecture}
\end{center}
\end{figure}

The Transformer~\cite{vaswani2017attention} architecture has been widely used for natural language processing. It has become the de-facto standard for many sequence-to-sequence tasks, such as machine translation. Transformer makes use of two computational blocks, an encoder and a decoder, both implemented by stacking multiple Transformer layers.
Transformer encoder layer consists of two consecutive layers, namely a self-attention layer followed by a position-wise feed-forward layer. Decoder adds third cross-attention layer, which attends over encoder output.
We sparsely scale Transformer with conditional computation by replacing every other feed-forward layer with a Position-wise Mixture of Experts (MoE) layer~\cite{shazeer2017outrageously} with a variant of top-2 gating in both the encoder and the decoder (Figure~\ref{fig:moe_architecture}). We vary the number of Transformer layers and the number of experts per MoE layer in order to scale the model capacity.

Each training example consists of a pair of sequences of subword tokens. Each token activates a sub-network of the MoE Transformer during both training and inference. The size of the sub-network is roughly independent of the number of experts per MoE Layer, allowing sublinear scaling of the computation cost as described in the previous section. Computation complexity is further analyzed in Section~\ref{sec:moela} and training performance in Section~\ref{sec:performance}.

\subsection{Position-wise Mixture-of-Experts Layer}~\label{sec:model-moe}

\newcommand{\combine}{\mathcal{G}}  % Combine weights
\newcommand{\gates}{g}    % raw gates

The Mixture-of-Experts (MoE) layer used in our model is based on \cite{shazeer2017outrageously} with variations in the sparse gating function and the auxiliary loss being used. A MoE layer for Transformer consists of $E$ feed-forward networks $\Expert_1 \dots \Expert_E$:

% hard to use generic top_k here
% hard to inline full gating definition with "dropping second expert"

% We can count non-zeros, but seems unnecessary
%   num = \sum_{e \in 1 dots E} (1 - \delta_{\combine,0})

% \begin{equation} compiler complains about nesting in equation, 
% \begin{align*} with & for alignment and * for no numbers
\begin{align}
    \combine_{s,E}           & = \Gate(x_s)                                         \\
    \Expert_e(x_s)         & = \WO_e \cdot \text{ReLU}(\WI_e \cdot x_s)                     \\
    y_s                    & = \sum^E_{e=1} \combine_{s,e} \cdot \Expert_e(x_s)      
\end{align}
% \end{equation}

where $x_s$ is the input token to the MoE layer, \WI and \WO being the input and output projection matrices for the feed-forward layer (an expert). Vector $\combine_{s,E}$ is computed by a gating network. $\combine_{s,E}$ has one non-negative for each expert, most of which are zeros meaning the token is not dispatched to that expert. The token is dispatched to a very small number of experts. We choose to let each token dispatched to at most two experts. The corresponding entries in $\combine_{s,E}$ are non-zeros, representing how much an expert contributes to the final network output. Every expert $\Expert_e$ applies to $x_s$ a fully-connected 2-layer network using ReLU~\cite{Nair2010RectifiedLU} activation function. The output of the MoE layer, $y_s$, is the weighted average of outputs from all the selected experts.

The gating function $\Gate(\cdot)$ is critical to the MoE layer, which is modeled by a softmax activation function to indicate the weights of each expert in processing incoming tokens. In other words, to indicate how good an expert is at processing the incoming token. Furthermore, the gating function must satisfy two goals:

\begin{itemize}
    \item \textbf{Balanced load} It is desirable that the MoE layer to sparsely activate the experts for a given token. A naive solution would be just to choose the top-$k$ experts according to the softmax probability distribution. However, it is known that this approach leads to load imbalance problem for training~\cite{shazeer2017outrageously}: most tokens seen during training would have been dispatched to a small number of experts, amassing a very large input buffer for only a few (busy) experts leaving other experts untrained, slowing down the training. Meanwhile many other experts do not get sufficiently trained at all. A better design of the gating function would distribute processing burden more evenly across all experts.

    \item \textbf{Efficiency at scale} It would be rather trivial to achieve a balanced load if the gating function is done sequentially. The computation cost for the gating function alone is at least $O(NE$) for all $N$ tokens in the input batch given $E$ experts. However, in our study, $N$ is in the order of millions and $E$ is in the order of thousands, a sequential implementation of the gating function would keep most of the computational resources idle most of the time. Therefore, we need an efficient parallel implementation of the gating function to leverage many devices.
\end{itemize}

We designed the following mechanisms in the gating function $\Gate(\cdot)$ to meet the above requirements (details illustrated in Algorithm~\ref{alg:gating}):

\begin{itemize}
    \item \textbf{Expert capacity}
     To ensure the load is balanced, we enforce that the number of tokens processed by one expert is below some uniform threshold, which we define as expert capacity. Assuming that the total number of tokens in a training batch is $N$, and each token is dispatched to at most two experts, then the expert capacity is set to be $O(N/E)$. $\Gate(\cdot)$ keeps a running counter $c_e$ for how many tokens are dispatched to an expert. When both experts selected by a token already exceed their capacity, the token is considered as an \textit{overflowed} token, where $\combine_{s,E}$ degenerates into a zero vector. Such tokens have their representation $x_s$ passed on to the next layer via residual connections.
     
    \item \textbf{Local group dispatching}     
    $\Gate(\cdot)$ partitions all tokens in a training batch evenly into $G$ groups, i.e., each group contains $S=N/G$ tokens. All groups are processed independently in parallel. Each group is given a fractional capacity of each expert, $2N/(G\cdot E)$. Each group ensures that at most this many tokens are dispatched to an expert. In this way, we can ensure that expert capacity is still enforced and the overall load is balanced.
    
    \item \textbf{Auxiliary loss}
    It is important that the gating function does not always choose the same few experts,
    as this would lead to a capacity overflow for only a few experts
    and under-utilization for the remaining ones.
    Following \cite{shazeer2017outrageously}, we define an auxiliary loss term $\ell_{aux}$ to enforce this constraint. It is added to the overall loss function of the model $\mathcal{L} = \ell_{nll} + k * \ell_{aux}$ with a constant multiplier $k$.
    %To balance expert utilization, reduce overflow and avoid winner-takes-all we add gating $\ell_{aux}$ %auxiliary losses to the training loss. \todo{DIMA to add a sentence on the meaning of aux loss}
    The particular form of the auxiliary loss term $\ell_{aux}$ in line (13) of algorithm~\ref{alg:gating} is motivated by the following consideration:
    the term $c_e/S$ represents the fraction of input routed to each expert,
    and we want to minimize mean square of $c_e/S$.
    But because $c_e$ is derived from top-2 operation and is not differentiable,
    we use the mean gates per expert $m_e$ as a differentiable approximation
    and replace $(c_e/S)^2$ with $m_e (c_e/S)$, which can now be optimized with gradient descent.

    \item \textbf{Random routing} Intuitively, because $y_s$ is a weighted average of what selected experts return, if the weight for the 2nd expert is very small, we can simply ignore the 2nd expert to conserve the overall expert capacity. Hence, in addition to respecting the expert capacity constraint, $\Gate(\cdot)$ dispatches to the 2nd-best expert with the probability proportional to its weight $g_2$.
\end{itemize}

% GSEC for ungrouped 128 devices, 128 experts/layer, and total batch of 1M tokens
% G=128 S=8192 E=128 C~=(GS/EG)=(S/E)=64
% 512 2048
\IncMargin{1em}
\SetNlSty{texttt}{(}{)}
\begin{algorithm}[t!] % The compiler isn't recognizing the float option 'H'. Include \usepackage{float} in your preamble to fix this.
\SetAlgoLined
\caption{Group-level top-2 gating with auxiliary loss}
\label{alg:gating}

% Need to reiterate that x_s is of size [M]?
\KwData{$x_S$, a group of tokens of size $S$}
\KwData{$C$, Expert capacity allocated to this group}

\KwResult{$\combine_{S,E}$, group combine weights}
\KwResult{$\ell_{aux}$, group auxiliary loss}

$c_{E} \gets 0$ \Comment{gating decisions per expert}

% GSE
$\gates_{S,E} \gets softmax(\WG \cdot x_S)$ \Comment{gates per token per expert, $\WG$ are trainable weights}

$m_{E} \gets \frac{1}{S} \sum^s_{s=1} \gates_{s,E}$ \Comment{mean gates per expert}

\newcommand{\foreverytoken}{$s \gets 1$ \KwTo $S$}

\For{\foreverytoken} {   %
    $\gates1, e1, \gates2, e2 = top\_2 (\gates_{s,E}) $    \Comment{top-2 gates and expert indices}
    
    $\gates1 \gets \gates1 / (\gates1+\gates2)$            \Comment{normalized $\gates1$}

    $c \gets c_{e1}$                    \Comment{position in $e1$ expert buffer}

    \If{$c_{e1} < C$} { 

        $\combine_{s,e1} \gets \gates1$   \Comment{$e1$ expert combine weight for $x_s$}
    }
    
    $c_{e1} \gets c + 1$                  \Comment{incrementing $e1$ expert decisions count}

}

\newcommand{\densityone}{\frac{c_e}{S}}

% loss computation uses capacity used by top-1 dispatches \frac{c_e}{S} and has to happen before second-best experts dispatch

% \Comment{auxiliary loss depends on a Expert capacity usage ratio by top-1 dispatches and mean gates per expert}
$\ell_{aux} = \frac{1}{E} \sum^E_{e=1} \densityone \cdot m_e$ 

\For{\foreverytoken} {   %
    $\gates1, e1, \gates2, e2 = top\_2 (\gates_{s,E}) $ \Comment{top-2 gates and expert indices}
    
    $\gates2 \gets  \gates2 / (\gates1 + \gates2)$ \Comment{normalized $\gates2$}
    
    $rnd \gets uniform(0, 1)$ \Comment{dispatch to second-best expert with probability $\propto 2 \cdot \gates2$}

    $c \gets c_{e2}$ \Comment{position in $e2$ expert buffer}

    \If{$c < C \land 2 \cdot \gates2 > rnd$} { 

        $\combine_{s,e2} \gets \gates2$ \Comment{$e2$ expert combine weight for $x_s$}
    }
    
    $c_{e2} \gets c + 1$    
}

\end{algorithm}

\section{Highly Parallel Implementation using \codename}
\label{sec:framwork}

This section describes the implementation of the model in Section~\ref{sec:model} that runs efficiently on a cluster of TPU devices.

The first step is to express the model in terms of linear algebra operations, in which our software stack (TensorFlow~\cite{abadi2016tensorflow}) and the hardware platform (TPU) are highly tailored and optimized. It is readily easy to code up most of the model in terms of linear algebra in the same way as the original Transformer. However, it requires some effort to express the MoE Layer, in particular $\Gate(\cdot)$ function presented in Algorithm~\ref{alg:gating} due to its sequential nature, and we describe the details in Section~\ref{sec:moela}.

Next, we annotate the linear algebra computation to express parallelism. Each tensor in the computation can be annotated for replication or distribution across a cluster of devices using sharding APIs in Section~\ref{sec:spmd_annotation}. Using sharding annotations enables separation of concerns between the model description and the efficient parallel implementation, and allows users to flexibly express diverse parallelization strategies. For example, (1) the attention layer is parallelized by splitting along the batch dimension and replicating its weights to all devices. On the other hand, (2) experts in the MoE layer are infeasible to be replicated in all the devices due to its sheer size and the only viable strategy is to shard experts into many devices. Furthermore, the whole model alternates between these two modes (1)-(2). Using annotations frees model developers from the system optimization efforts and avoids baking the parallel implementation and low-level details into the model code.

Finally, the compiler infrastructure takes a (partially) annotated linear algebra computation and produces an efficient parallel program that scales to thousands of devices. As will be described in Section~\ref{sec:partitioner}, the compiler applies SPMD (Single Program Multiple Data) partitioning transformation to express per-device computation, inserts necessary cross-device communication, handles irregular patterns such as uneven partitions, and finally generates a single program to be launched on all devices for parallel execution.

\subsection{Positions-wise Mixture-of-Expert Layer Expressed in Linear Algebra}
\label{sec:moela}

% The Power of Abstraction
% Inconvenience stems from partitioning implementation  from  model description should be separated from the partitioning implementation and optimization. 

Our model implementation (Algorithm~\ref{alg:moe}) views the whole accelerator cluster as a single device and expresses its core mathematical algorithm in a few tensor operations independent of the concrete setup of the cluster. Einstein summation notation~\cite{einstein1923grundlage} (i.e., tf.einsum) is a powerful construct to concisely express the model and we use it extensively in our implementation.
The softmax gates computation is trivially expressed by one einsum followed by the softmax function. Dispatching of inputs to selected experts is expressed by a single einsum between the dispatching mask and the input. All $\Expert_e$ weights are combined into single 3-D tensors \texttt{wi} amd \texttt{wo} and the computation by $\Expert_1 \dots \Expert_E$ is expressed using 3 operators (two einsum and one relu). Finally, taking weighted average of all experts output into the final output is expressed in another einsum.

\texttt{Top2Gating} in Algorithm~\ref{alg:moe} computes the union of all group-local $\combine_{S,E}$ described in Algorithm~\ref{alg:gating}. \texttt{combine\_weights} is a 4-D tensor with shape \texttt{[G, S, E, C]}.
The value \texttt{combine\_weights[g, s, e, c]} is non-zero
when the input token $s$ in group $g$ 
is sent to the input buffer
of expert $e$ at buffer position $c$.
%
%It is a dense representation for token routing weights to %all expert output positions. 
For a specific \texttt{g} and \texttt{s}, a slice \texttt{combine\_weight[g, s, :, :]} contains at most two non-zero vaules.
%Let us assume that expert \texttt{e1} and expert \texttt{e2} %are chosen for the token \texttt{s}.  The token occupies the %\texttt{c1}-th and \texttt{c2}-th slots of expert \texttt{e1} %and \texttt{e2} respectively. \texttt{combine\_weights[g, s, %:, :]} are all zeros except \texttt{combine\_weights[g, s, %e1, c1]} and \texttt{combine\_weights[g, s, e2, c2]}.
Binary \texttt{dispatch\_mask} is produced from \texttt{combine\_weights} by simply setting all non-zero values to 1.

\newcommand{\Sharded}[1]{\textcolor{red}{\underline{#1}}}
\newcommand{\GG}{\Sharded{G}}
\newcommand{\EE}{\Sharded{E}}

\begin{algorithm} % enter the algorithm environment
\caption{Forward pass of the Positions-wise MoE layer. The underscored letter (e.g., \GG\ and \EE) indicates the dimension along which a tensor will be partitioned.} % give the algorithm a caption

\label{alg:moe} % and a label for \ref{} commands later in the document
% dispatch_mask is binary

\begin{lstlisting}[escapeinside={&}{&},numbers=left,    stepnumber=1]
gates = softmax(einsum("&\GG&SM,ME->&\GG&SE", inputs, wg))
combine_weights, dispatch_mask = Top2Gating(gates)
dispatched_expert_inputs = einsum(
    "&\GG&SEC,&\GG&SM->&\EE&GCM", dispatch_mask, reshaped_inputs)
h = einsum("&\EE&GCM,&\EE&MH->&\EE&GCH", dispatched_expert_inputs, wi)
h = relu(h)
expert_outputs = einsum("&\EE&GCH,&\EE&HM->&\GG&ECM", h, wo)
outputs = einsum(
    "&\GG&SEC,&\GG&ECM->&\GG&SM", combine_weights, expert_outputs)
\end{lstlisting}
\end{algorithm}

We need to choose the number of groups $G$ and the number of experts $E$ properly so that the algorithm can scale to a cluster with $D$ devices. It is worthwhile to analyze its overall computation complexity (the total number of floating point operations) for a training step given a training batch of $N$ tokens.

We analyze Algorithm~\ref{alg:moe} computation complexity scaling with number the of devices $D$  with the following assumptions:
\textit{a)} number of tokens per device $\frac{N}{D}=O(1)$ is constant\footnote{This is oftentimes necessary in practice to avoid overflowing device memory.};
\textit{b)} $G=O(D)$, $S=O(1)$ and $N=O(GS)=O(D)$;
\textit{c)} $M=O(1)$, $H=O(1)$;
\textit{d)} $E=O(D)$; and 
\textit{e)} $C=O(\frac{2S}{E})=O(\frac{1}{D}), D<S$ and is a positive integer\footnote{Scaling $D>S$ would require different use of fractional expert capacity.}
% \begin{enumerate*}[label=\textit{\alph*}), 
%       itemjoin={{; }},itemjoin*={{; and }}]
% \item number of tokens per device $\frac{N}{D}=O(1)$ is constant\footnote{This is oftentimes necessary in practice to avoid overflowing device memory.}
% \item $G=O(D)$, $S=O(1)$ and $N=O(GS)=O(D)$
% \item $M=O(1)$, $H=O(1)$
% \item $E=O(D)$ 
% \item $C=O(\frac{2S}{E})=O(\frac{1}{D}), D<S$ and is a positive integer\footnote{Scaling $D>S$ would require different use of fractional expert capacity.}
% \end{enumerate*}
.

The total number of floating point operations $FLOPS$ in Algorithm~\ref{alg:moe}:
\begin{alignat*}{4}
          & \\
          &FLOPS_{\text{Softmax}}  
          &+& FLOPS_{\text{Top2Gating}}  
          &+& FLOPS_{\text{Dispatch|Combine}}
          &+& FLOPS_{\Expert} =\\
          &O(GSME)
          &+& O(GSEC)
          &+& O(GSMEC)
          &+& O(EGCHM) =\\
          &O(D \cdot 1 \cdot 1 \cdot D)
          &+& O(D \cdot 1 \cdot D \cdot \frac{1}{D})
          &+& O(D \cdot 1 \cdot 1 \cdot D \cdot \frac{1}{D})
          &+& O(D \cdot D \cdot \frac{1}{D} \cdot 1 \cdot 1) =\\  
          &O(D^2)
          &+& O(D)
          &+& O(D)
          &+& O(D)
\end{alignat*}

% \todo{if you think the equation here is cluttered, another alternative for display purposes:\\
% \begin{matrix}\\
% Dispatch\vert Combine
% \\ 
% \overbrace{O(GSMEC)}  + \\
% \end{matrix}
% }

and consequently per-device $FLOPS/D = O(D) + O(1) + O(1) + O(1)$. Per-device softmax complexity $FLOPS_\text{softmax}/D=O(D)$ is linear in number of devices, but in practice is dominated by other terms since $D << H$ and $D < S$. As a result $FLOPS/D$ could be considered $O(1)$, satisfying sublinear scaling design requirements. Section~\ref{sec:performance} verifies this analysis empirically.

In addition to the computation cost, we have non-constant cross-device communication cost, but it grows at a modest rate $O(\sqrt{D})$ when we increase $D$ (Section~\ref{sec:performance}).

\subsection{\codename Annotation API for Parallel Execution}
\label{sec:spmd_annotation}

Due to the daunting size and computation demand of tensors in Algorithm~\ref{alg:gating}, we have to parallelize the algorithm over many devices. An immediate solution of how to shard each tensor in the algorithm is illustrated by underscored letters in Algorithm~\ref{alg:moe}. The \emph{sharding} API in \codename allows us to annotate tensors in the program to selectively specify how they should be partitioned. This information is propagated to the compiler so that the compiler can automatically apply transformations for parallel execution. We use the following APIs in TensorFlow/Lingvo~\cite{shen2019lingvo} in our work.

\begin{itemize}
    \item \textbf{replicate(tensor)} annotates \texttt{tensor} to be replicated across partitions, and returns the annotated tensor. This is often used for the non-MoE layers in our model to replicate the weights.
    \item \textbf{split(tensor, split\_dimension, num\_partitions)} annotates \texttt{tensor} to be partitioned along \texttt{split\_dimension}, and returns the annotated tensor. Partition $i$ is placed on the $i$'th device, and \texttt{num\_partitions} must not exceed the number of devices on the system.
    \item \textbf{shard(tensor, device\_assignment)} generalizes \texttt{split()} to allow partitioning multiple dimensions and specifying the placement of each partition. Appendix~\ref{sec:appendix-shard-api} describes this API with more details.
\end{itemize}

Note that the invocations to \texttt{split} or \texttt{shard} only adds annotations and does not change the logical shape in the user program. The user still works with full shapes and does not need to worry about issues like uneven partitioning.

\codename is general in the sense that the simple APIs apply to all dimensions in the same way. The sharded dimensions could include batch (data-parallelism), feature, expert, and even spatial dimensions in image models, depending on the use cases. Also, since the sharding annotation is per tensor, different parts of the model can be partitioned in different ways. This flexibility enables us to partition the giant MoE weights and switch partition modes between MoE and non-MoE layers, as well as uses cases beyond this paper, e.g., spatial partitioning of large images~\cite{spatial-partitioning} (Appendix~\ref{sec:appendix-spmd-conv}).

With the above sharding APIs, we can express the sharding strategy shown in Algorithm~\ref{alg:moe} as below. The input tensor is split along the first dimension and the gating weight tensor is replicated. After computing the dispatched expert inputs, we apply \texttt{split} to change the sharding from the group ($G$) dimension to the expert ($E$) dimension. $D$ is device count.
\newcommand{\Hilight}{\makebox[0pt][l]{\color[HTML]{bef5cb}\rule[-4pt]{0.99\linewidth}{14pt}}}

\begin{lstlisting}[escapeinside={&}{&},numbers=left,    stepnumber=1,escapechar=\%]
  # Partition inputs along group (G) dim. 
%\Hilight%+ inputs = split(inputs, 0, D)
  # Replicate the gating weights
%\Hilight%+ wg = replicate(wg)
  gates = softmax(einsum("GSM,ME->GSE", inputs, wg))
  combine_weights, dispatch_mask = Top2Gating(gating_logits)
  dispatched_expert_inputs = einsum(
    "GSEC,GSM->EGCM", dispatch_mask, reshaped_inputs)
  # Partition dispatched inputs along expert (E) dim.
%\Hilight%+ dispatched_expert_inputs = split(dispatched_expert_inputs, 0, D)
  h = einsum("EGCM,EMH->EGCH", dispatched_expert_inputs, wi)
  ...
\end{lstlisting}

\paragraph{Per-tensor sharding assignment} As shown in the example above, users are not required to annotate every tensor in the program. Annotations are typically only required on a few important operators like Einsums in our model and the compiler uses its own heuristics to infer sharding for the rest of the tensors~\footnote{It is also important for the compiler to infer missing shardings since the backpropagation computation is often automatically generated by the frontend framework and users don't have access to those tensors.}. For example, since the input tensor is partitioned along $G$ and the weight tensor is replicated, the compiler chooses to partition the einsum output along the same $G$ dimension (Line 5). Similarly, since both inputs are partitioned along the $G$ dimension for the input dispatch einsum (Line 7), the output sharding is inferred to be split along the $G$ dimension, and then we add the \texttt{split} annotation on the output to reshard along the $E$ dimension. Some annotations in the above example could also be determined by the compiler (e.g., \texttt{replicate(wg)}) but it is recommended to annotate the initial input and final output tensors of the computation.

The compiler currently uses an iterative data-flow analysis to propagate sharding information from an operator to its neighbors (operands and users), starting from the user-annotated operators. The analysis tries to minimize the chance of resharding by aligning the sharding decisions of adjacent operators. There could be other approaches such as integer programming or machine-learning methods, but improving the automatic sharding assignment is not the focus of this paper and we leave it as future work.

\paragraph{Mixing manual and automatic sharding} Automatic partitioning with sharding annotations is often enough for common cases, but \codename also has the flexibility to allow mixing manually partitioned operators with auto-partitioned operators. This provides users with more controls on how operators are partitioned, and one example is that the user has more run-time knowledge beyond the operators' semantics. For example, neither XLA's nor TensorFlow's \texttt{Gather} operator definition conveys information about the index bounds for different ranges in the input, but the user might know that a specific \texttt{Gather} operator shuffles data only within each partition. In this case, the user can trivially partition the operator by simply shrinking the dimension size and performing a local \texttt{Gather}; otherwise, the compiler would need to be conservative about the index range and add unnecessary communication overhead. For example, the dispatching \texttt{Einsum} (Line 3) in Algorithm~\ref{alg:moe} in Algorithm~\ref{alg:moe}, which uses an one-hot matrix to dispatch inputs, can be alternatively implemented with a \texttt{Gather} operator using trivial manual partitioning, while the rest of the model is partitioned automatically. Below is the pseudocode illustrating this use case.

\begin{lstlisting}[escapeinside={&}{&},numbers=left,    stepnumber=1,escapechar=\%]
# input has shape [G, S, M]. split() does not change logical shape.
input = split(input, 0, num_devices)
# s_indices has shape [E, G, C, 1]. Values: indices to S in input.
s_indices = split(s_indices, 1, num_devices)

# Begin manual partitioning.
# partitioned_input has shape [G/num_devices, S, M]
%\Hilight%partitioned_input = auto_to_manual_spmd_partition(input)
# partitioned_s_indices has shape [E, G/num_devices, C, 1]
%\Hilight%partitioned_s_indices = auto_to_manual_spmd_partition(s_indices)
# Concat with G indices in partitioned_input: Iota on G dimension.
partitioned_gs_indices = concat(
    iota([E, G/num_devices, C, 1], 1), partitioned_s_indices, 3)
# partitioned_data has shape [E, G/num_devices, C, M]
partitioned_data = gather(
    partitioned_input, partitioned_gs_indices)

# Switch back to auto partitioning.
# data has shape [E, G, C, M]
%\Hilight%data = manual_to_auto_spmd_partition(partitioned_data)
...
\end{lstlisting}

%While this particular case could also be solved by extending the definition of \texttt{Gather} to add a parallel dimension (e.g., segmented gather), or by performing additional static analysis on index ranges, the mixing mechanism can useful in general for power users who want to control how an operator is partitioned (not only controlling how tensors are partitioned using annotations).

\subsection{The XLA SPMD Partitioner for \codename}\label{sec:partitioner}
This section describes the compiler infrastructure that automatically partitions a computation graph based on sharding annotations. Sharding annotations inform the compiler about how each tensor should be distributed across devices. The SPMD (Single Program Multiple Data) partitioner (or ``partitioner'' for simplicity) is a compiler component that transforms a computation graph into a single program to be executed on all devices in parallel. This makes the compilation time near constant regardless of the number of partitions, which allows us to scale to thousands of partitions.~\footnote{An alternative is MPMD (Multiple Program Multiple Data), which does not scale as shown in Figure~\ref{fig:spmd_mpmd_comparison}.}

We implemented the partitioner in the XLA compiler~\cite{xla}. Multiple frontend frameworks including TensorFlow, JAX, PyTorch and Julia already have lowering logic to transform their graph representation to XLA HLO graph. XLA also has a much smaller set of operators compared to popular frontend frameworks like TensorFlow, which reduces the burden of implementing a partitioner without harming generality, because the existing lowering from frontends performs the heavy-lifting to make it expressive. Although we developed the infrastructure in XLA, the techniques we describe here can be applied to intermediate representations in other machine learning frameworks (e.g., ONNX~\cite{onnx}, TVM Relay~\cite{roesch2018relay}, Glow IR~\cite{rotem2018glow}).

% Input is a dataflow graph, the partitioner visits each node to transform.
XLA models a computation as a dataflow graph where nodes are operators and edges are tensors flowing between operators. The core of the partitioner is per-operation handling that transforms a full-sized operator into a partition-sized operator according to the sharding specified on the input and output. When a computation is partitioned, various patterns of cross-device data transfers are introduced. In order to maximize the performance at large scale, it is essential to define a core set of communication primitives and optimize those for the target platform. 

%While some operations (e.g., elementwise) are trivial to support, we discuss several common cases where cross-partition communications are required.

\subsubsection{Communication Primitives}
Since the partitioner forces all the devices to run the same program, the communication patterns are also regular and XLA defines a set of collective operators that perform MPI-style communications~\cite{mpi2.2}. We list the common communication primitives we use in the SPMD partitioner below.

\paragraph{CollectivePermute} This operator specifies a list of source-destination pairs, and the input data of a source is sent to the corresponding destination. It is used in two places: changing a sharded tensor's device order among partitions, and halo exchange as discussed later in this section.

\paragraph{AllGather} This operator concatenates tensors from all participants following a specified order. It is used to change a sharded tensor to a replicated tensor.

\paragraph{AllReduce} This operator performs elementwise reduction (e.g., summation) over the inputs from all participants. It is used to combine partially reduced intermediate tensors from different partitions. In a TPU device network, \texttt{AllReduce} has a constant cost when the number of partition grows (Section~\ref{sec:perf-scalability}). It is also a commonly used primitive with efficient implementation in other types of network topology~\cite{cho2019blueconnect}.

\paragraph{AllToAll} This operator logically splits the input of each participant along one dimension, then sends each piece to a different participant. On receiving data pieces from others, each participant concatenates the pieces to produce its result. It is used to reshard a sharded tensor from one dimension to another dimension. \texttt{AllToAll} is an efficient way for such resharding in a TPU device network, where its cost increases sublinearly when the number of partitions grows (Section~\ref{sec:perf-scalability}).

\subsubsection{Per-Operator SPMD Partitioning}
\label{sec:op-partitioner}
The core of the partitioner is the per-operator transformation from a full-sized operator into a partition-sized operator according to the specified sharding. While some operators (e.g., elementwise) are trivial to support, we discuss several common cases where cross-partition communications are required.

There are a few important technical challenges in general cases, which we will cover in Section~\ref{sec:spmd-general-cases}. To keep the discussion more relevant to the MoE model, this section focuses on \texttt{Einsum} partitioning to illustrate a few communication patterns. And to keep it simple for now, we assume that all tensors are evenly partitioned, which means the size of the dimension to partitition is a multiple of the partition count.

\paragraph{Einsum Case Study}
\texttt{Einsum} is the most critical operator in implementing the MoE model. They are represented as a \texttt{Dot} operation in XLA HLO, where each operand (LHS or RHS) consists of three types of dimensions:
\begin{itemize}
    \item \textbf{Batch dimensions} are the embarrassingly parallel dimensions. The same set of batch dimensions must exist in all of LHS, RHS and the output, and each element in the output only depends on the corresponding batch in LHS and RHS.
    \item \textbf{Contracting dimensions} only exist in the operands. LHS and RHS must have the same set of contracting dimensions, and they are summed up and collapsed in the output.
    \item \textbf{Non-contracting dimensions} are also parallel dimensions that exist in one of the operands and the output. Each of LHS and RHS has its own set of non-contracting dimensions, which are inherited by the output.
\end{itemize}

Sharding propagation prioritizes choosing the same sharding on batch dimensions of LHS, RHS and output, because that would avoid any cross-partition communication. However, that is not always possible, and we need cross-partition communication in the following three cases.

\begin{figure}[t!]
\begin{center}
\subfloat[A partitioned \texttt{Einsum} operator. Colored letters ($G$ and $E$) represent the partitioned dimension of each tensor. The partitioner decides to first execute a batch-parallel \texttt{Einsum} along the $G$ dimension, then reshard the result to the $E$ dimension.]{
\includegraphics[width=0.7\textwidth]{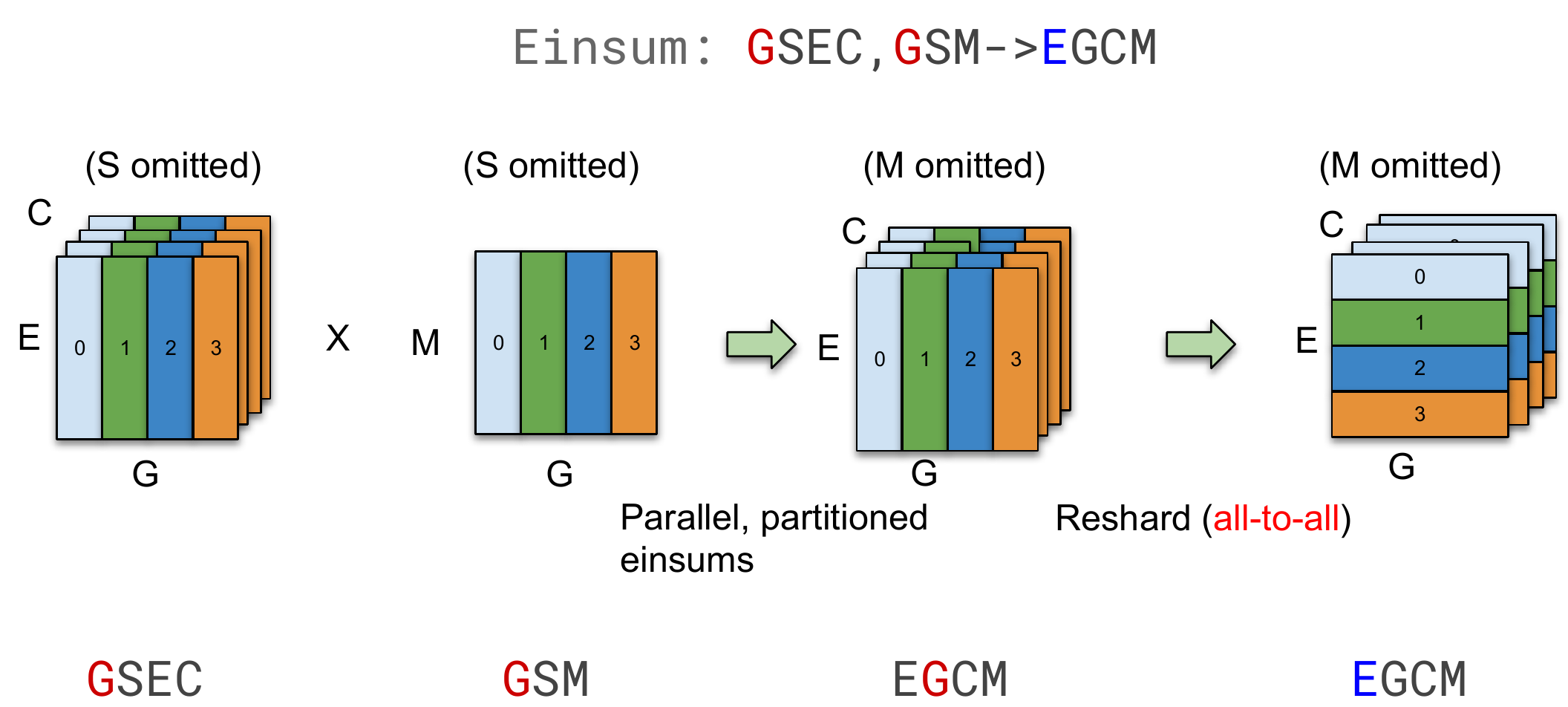}
\label{fig:einsum_example}
}

\subfloat[A simple \texttt{Einsum} (\texttt{Matmul}) partitioned on the contracting dimension.]{
\includegraphics[width=0.7\textwidth]{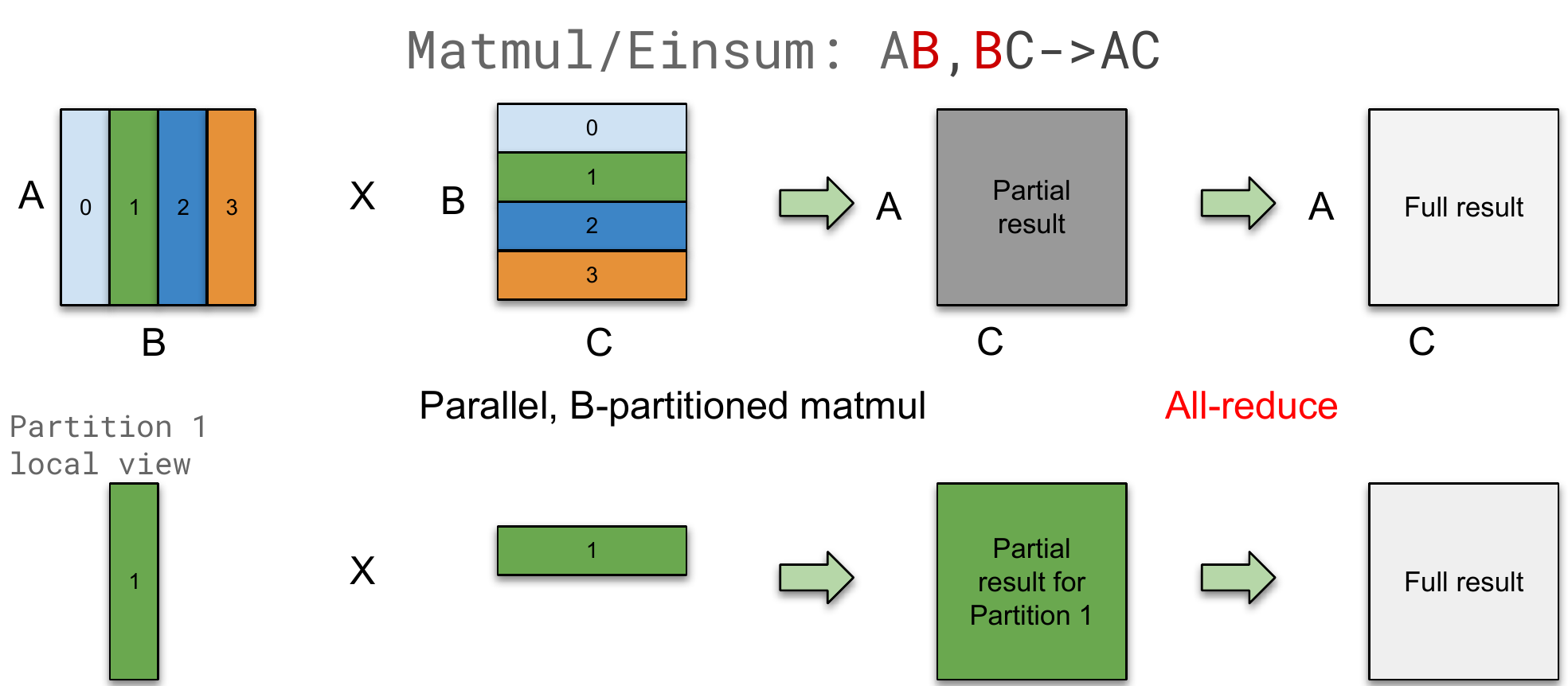}
\label{fig:dot_example}
}

\subfloat[An \texttt{Einsum} (\texttt{Matmul}) where we use collective-permute in a loop to compute one slice at a time. There is no full-sized tensor during the entire process.]{
\includegraphics[width=0.7\textwidth]{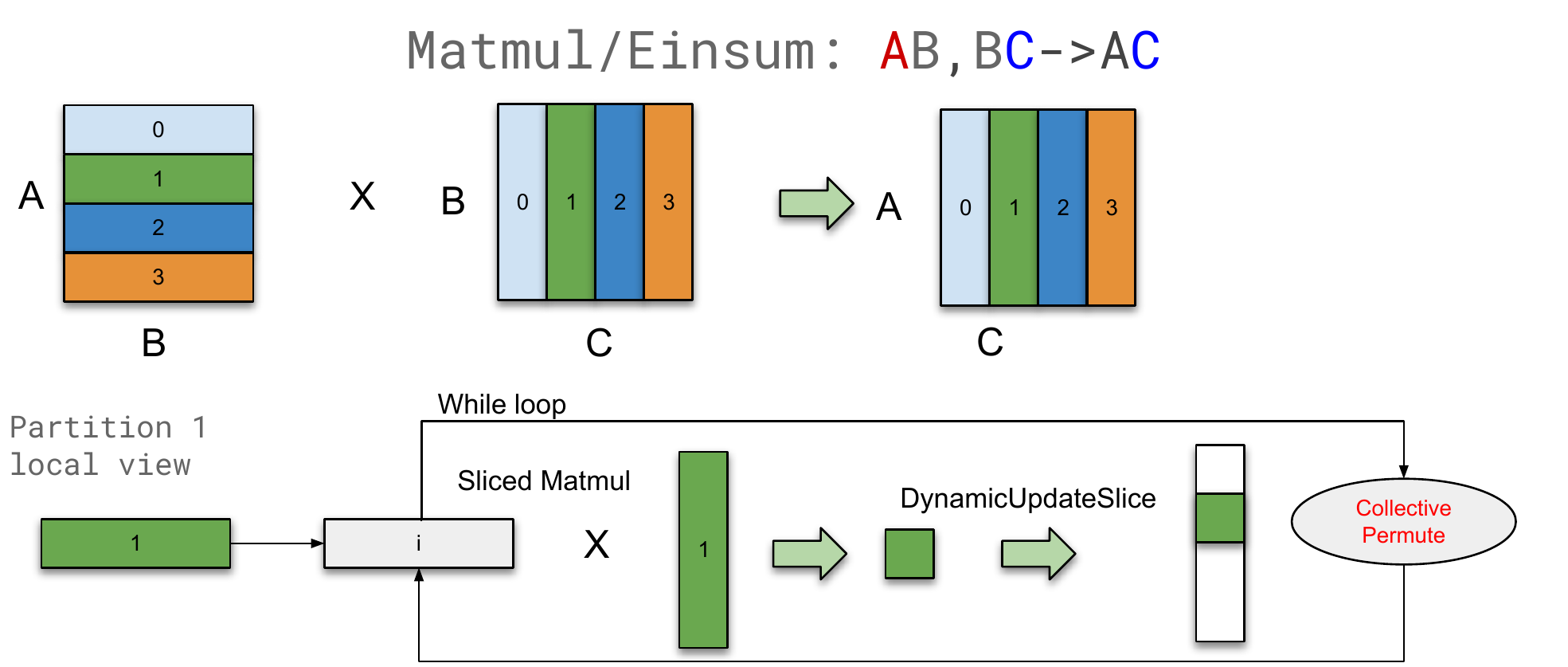}
\label{fig:cannon_example}
}
\caption{Examples of \texttt{Einsum} partitioning with cross-device communication.}
\end{center}
\end{figure}

\begin{itemize}
    \item \textbf{Resharding.} In the MoE model we built, the expert dispatching logic (Line 3 in Algorithm~\ref{alg:moe}) requires switching the partitioned dimension after an \texttt{Einsum}. Since resharding is efficient (Section~\ref{sec:perf-scalability}) with \texttt{AllToAll}, we first execute the \texttt{Einsum} locally, then reshard it to the desired dimension, as shown in Figure~\ref{fig:einsum_example}.
    
    \item \textbf{Accumulating partial results.} If the inputs are partitioned along contracting dimensions, the local result is partial and we need to use an \texttt{AllReduce} to combine them and produce the final result, as shown in Figure~\ref{fig:dot_example}.
    
    \item \textbf{Slicing in a loop.} For certain scenarios, we also implemented an algorithm similar to Cannon's algorithm~\cite{cannon1969}, in order to limit the size of tensors on each partition. For example, if both operands are partitioned on a non-contracting dimension, we cannot compute the local \texttt{Einsum} directly since operands have different non-contracting dimensions. Replicating one of the operands would not cause redundant computation, but it requires the replicated operand to fit in device memory. Therefore, if the size of the operand is too large, we instead keep both operands partitioned and use a loop to iterate over each slice of the result, and use \texttt{CollectivePermute} to communicate the input slices (Figure~\ref{fig:cannon_example}).
\end{itemize}

\subsubsection{Supporting a Complete Set of Operators} %~
\label{sec:spmd-general-cases}

We solved several additional challenges to enable the SPMD partitioner to support a complete set of operators without extra constraints of tensor shapes or operator configurations.
These challenges often involve asymmetric compute or communication patterns between partitions, which are particularly hard to express in SPMD, since the single program needs to be general enough for all partitions. We cannot simply create many branches in the single program based on the run-time device ID, because that would lead to an explosion in program size.
%without adding constraints to shapes or operator configurations.

% The examples in Section~\ref{sec:op-partitioner} are relatively simple for the SPMD partitioner, because with the right communication primitive, all partitions have uniform behavior. However, it is also common that a tensor may not be evenly partitioned, and/or different partitions' computations may be different enough to express trivially.

% The design choice of using a single program is key to compiler \emph{scalability}, but comes with challenges to keep it \emph{general} for non-uniform cases. With multiple programs, each program could be specialized to one partition; but with a single program, it has to be general enough to express computations on all partitions. We cannot simply create many branches in the single program based on the run-time device ID, because that would lead to an explosion in program size. The rest of the section discusses a set of novel techniques we use to address these challenges.

% \paragraph{Building blocks for non-uniformity}
% We use a small set of XLA operators that enable non-uniform behavior and data accesses between partitions while still preserving the one-program representation. These include dynamic offset calculation (via \texttt{PartitionId} and scalar integer operations), data alignment (via \texttt{DynamicSlice} and \texttt{Pad}), as well as value masking (via \texttt{Compare} on indices and \texttt{Select}).

\paragraph{Static shapes and uneven partitioning}
XLA requires tensor shapes to be static.~\footnote{The limited dynamism in the intermediate representation is often necessary to efficiently target accelerators.}
However, when a computation is partitioned, it's not always the case that all partitions have the same input/output shapes, because dimensions may not be evenly divisible by the number of partitions.
In those cases, the size of the shape is rounded up to the next multiple of partition count, and the data in that padded region can be arbitrary.

When computing an operator, we may need to fill in a known value to the padded region for correctness. For example, if we need to partition an \texttt{Reduce-Add} operator, the identity value of zero needs to be used.
Consider an example where the partitioned dimension (15) cannot be divided into 2 (partition count), so Partition 1 has one more column than needed. We create an \texttt{Iota} operator of range [0, 8), add the partition offset (calculated from $PartitionId \times 8$), and compare with the full shape offset (15). Based on the predicate value, we select either from the operand or from zero, and the result is the masked operand.

\paragraph{Static operator configurations} XLA operators have static configurations, like the padding, stride, and dilation defined in \texttt{Convolution}. However, different partitions may not execute with the same operator configuration. E.g., for a \texttt{Convolution}, the left-most partition applies padding to its left while the right-most partition applies padding to its right. In such cases, the partitioner may choose configurations that make some partitions to produce slightly more data than needed, then slice out the the irrelevant parts. Appendix~\ref{sec:appendix-spmd-conv} discusses examples for \texttt{Convolution} and similar operators.

\paragraph{Halo exchange} Certain operators have a communication pattern which involves partial data exchange with neighboring partitions, which we call \emph{halo exchange}. We use the \texttt{CollectivePermute} operator to exchange halo data between partitions.

\begin{figure}[t!]
\subfloat[Convolution]{
 \includegraphics[width=0.32\textwidth]{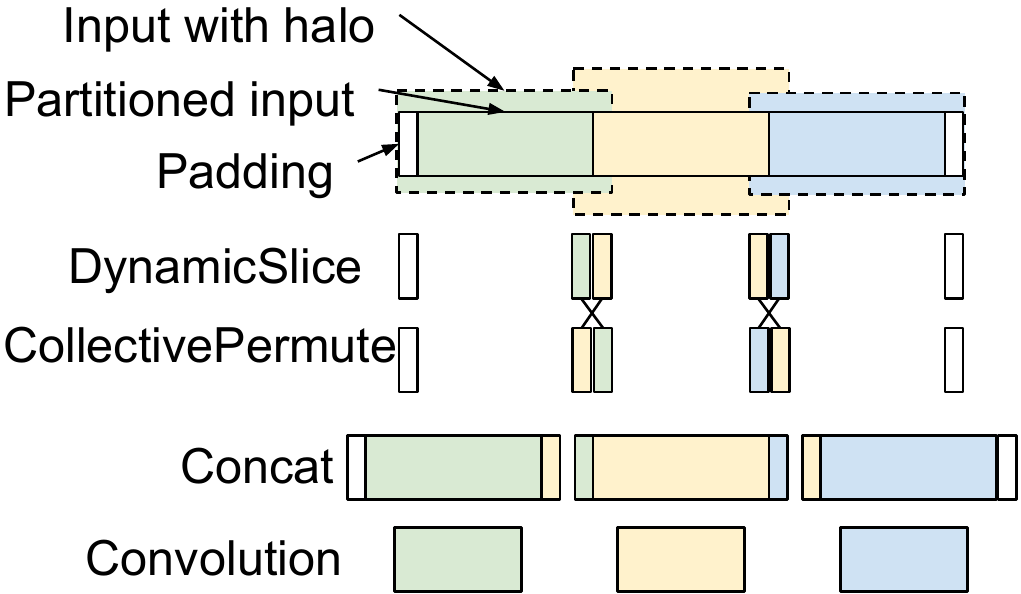}
 \label{fig:halo_exchange_conv}
}
\subfloat[Pad]{
 \includegraphics[width=0.32\textwidth]{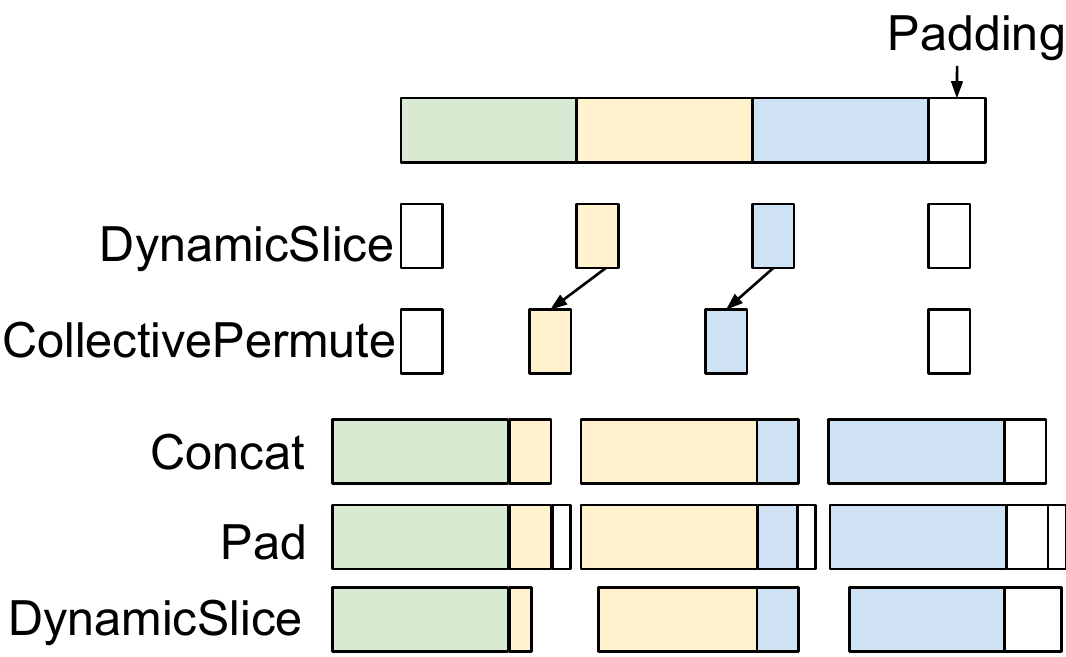}
 \label{fig:halo_exchange_pad}
}
\subfloat[Reshape with unevenly partitioned input and evenly partitioned output]{
 \includegraphics[width=0.32\textwidth]{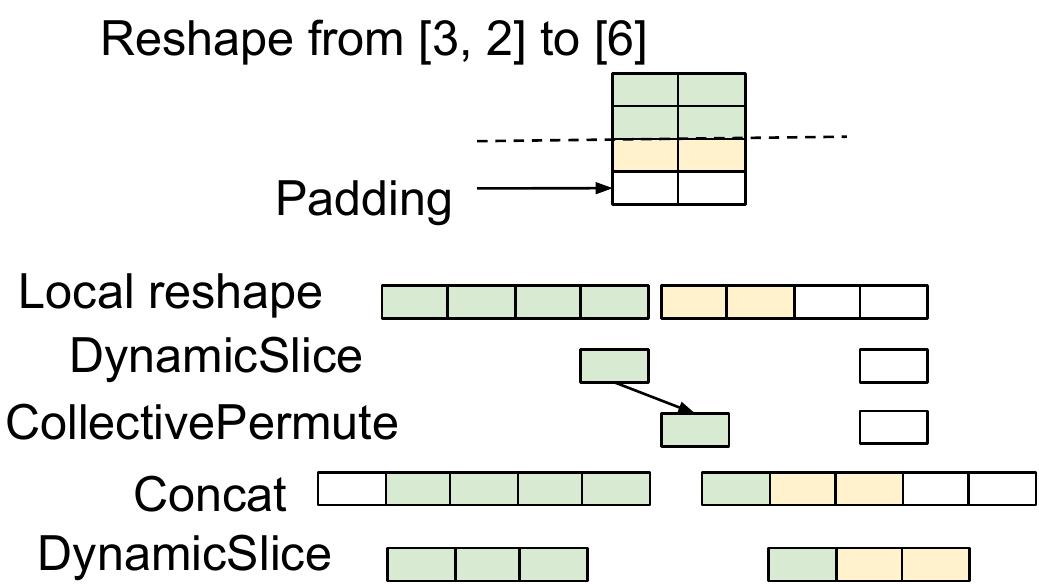}
 \label{fig:halo_exchange_reshape}
}
\caption{Halo exchange examples.}
\end{figure}

The most typical use case of halo exchange is for partitinoning window-based operators (e.g., \texttt{Convolution}, \texttt{ReduceWindow}), because neighboring partitions may require overlapping input data (Figure~\ref{fig:halo_exchange_conv}). In practice, halo-exchange for these operator often needs to be coupled with proper padding, slicing, and masking due to advanced use of window configurations (dilation, stride, and padding), as well as uneven halo sizes. We describe various scenarios in Appendix~\ref{sec:appendix-spmd-conv}.

Another use of halo exchange is for data formatting operators that change the size of the shape. For example, after a \texttt{Slice} or \texttt{Pad} operator, the shape of the tensor changes, and so do the boundaries between partitions. This requires us to realign the data on different partitions, which can be handled as a form of halo exchange (Figure~\ref{fig:halo_exchange_pad}).

Other data formatting operators, although logically not changing the size of the shape, may also need halo exchange, specifically due to the static shape constraint and uneven partitioning. For example, the \texttt{Reverse} operator reverses the order of elements in a tensor, but if it is partitioned unevenly, we need to shift data across partitions to keep the padding logically to the right of the result tensor. Another example is \texttt{Reshape}. Consider reshaping a tensor from [3, 2] to [6], where the input is unevenly partitioned in 2 ways on the first dimension (partition shape [2, 2]), and the output is also partitioned in 2 ways (partition shape [3]). There is padding on the input due to uneven partitioning, but after \texttt{Reshape}, the output tensor no longer has padding; as a result, halo exchange is required in a similar way to \texttt{Slice} (Figure~\ref{fig:halo_exchange_reshape}).

\paragraph{Compiler optimizations}
The SPMD partitioner creates various data formatting operators in order to perform slicing, padding, concatenation, masking and halo exchange.
%Since some of these operators wouldn't be needed if we compiled separate programs for each partition (i.e., Multiple Program Multiple Data), the runtime overhead from those ops may limit the benefits of partitioning.
To address the issue, we leverage XLA's fusion capabilities on TPU, as well as code motion optimizations for slicing and padding, to largely hide the overhead of data formatting. As a result, the run-time overhead is typically negligible, even for convolutional networks where masking and padding are heavily used.
\section{Massively Multilingual, Massive Machine Translation (M4)}
\label{sec:experiments}

\subsection{Multilingual translation}
\label{sec:mnmt}
% \todo{Summary: a) we chose MT; b) what is multilingual MT; c) our work towards universal MT; d) real-world impact; missing: why it's a good task to use? highlight multi-tasking?}
 We chose multilingual neural machine translation (MT)~\cite{Firat_2016,Johnson_2017,DBLP:journals/corr/abs-1903-00089}  to validate our design for efficient training with \codename. 
 Multilingual MT, which is an inherently multi-task learning problem, aims at building a single neural network for the goal of translating multiple language pairs simultaneously. 
 This extends our line of work~\cite{gpipe19,arivazhagan2019massively,shazeer2017outrageously} towards a universal machine translation model~\cite{translate2019m4}, i.e. a single  model that can translate between more than hundred languages, in all domains. 
Such massively multilingual translation models are not only convenient for stress testing models at scale, but also shown to be practically impactful in real-world production systems ~\cite{translate2020quality}.

% \todo{Summary: a) multilingual MT is improving \textit{high-resource} + \textit{low-resource} b) improving \textit{low-resource} via positive transfer (sharding target language model with \textit{high resource} c) \textit{high-resource} compete between each other for capacity and suffer from negative transfer d) notion of \textit{capacity bottleneck}}
In massively multilingual MT, there are two criteria that define success in terms of the model quality, 1) improvements attained on languages that have large amounts of training data (high resourced), and 2) improvements for languages with limited data (low-resource). As the number of language pairs (tasks) to be modeled within a single translation model increases, \textit{positive language transfer} \cite{baldwin1988transfer} starts to deliver large gains for low-resource  languages. Given the number of languages considered, M4 has a clear advantage on improving the low-resource tasks. On the contrary, for high-resource languages
the increased number of tasks limits per-task capacity within the model, resulting in lower translation quality compared to a models trained on a single
language pair.
This \textit{capacity bottleneck} for high resourced languages can be relaxed by increasing the model size to massive scale in order to satisfy the need for additional capacity \cite{arivazhagan2019massively,gpipe19}. 

% \todo{Summary: a) increase positive transfer via multilinguality; b) reduce negative transfer for high-resource by mitigating capacity bottleneck; c) search space in model design for multi-tasking grows rapidly with number of tasks. Maybe merge with previous paragraph?}
Massively multilingual, massive MT consequently aims at striking a balance between increasing \textit{positive transfer} by massive multilinguality and mitigating the \textit{capacity bottleneck} by massive scaling. While doing so, scaling the model size and the number of languages considered have to be coupled with a convenient neural network architecture. In order to amplify the \textit{positive transfer} and reduce the \textit{negative transfer}\footnote{Negative transfer is the notion of sharing the model capacity by unrelated tasks which in return hurts the quality of such \textit{interfering} tasks.}, one can naturally design a model architecture that harbours shared components across languages (shared sub-networks), along with some language specific ones (unshared, language specific sub-networks). However, the search space in model design (deciding on what to share) grows rapidly as the number of languages increase, making heuristic-based search for a suitable architecture impractical. Thereupon the need for approaches based on learning the wiring pattern of the neural networks from the data emerge as scalable and practical way forward. 

% \todo{Summary: a) conditional computation with GShard; b) study convergence; model quality, training efficiency; c) efficient multi-tasking with MoE with no prior knowledge about the need for tasks capacity and their interference.}
In this section, we advocate how conditional computation \cite{bengio2013estimating,davis2013lowrank} with sparsely gated mixture of experts \cite{shazeer2017outrageously} fits into the above detailed desiderata and show its efficacy by \textbf{scaling neural machine translation models beyond 1 trillion parameters}, while keeping the training time of such massive networks practical. E.g. a 600B GShard model for M4 can process 1T tokens\footnote{Source side tokens after sub-word segmentation.} in 250k training steps in under 4 days. We experiment with increasing the model capacity by adding more and more experts into the model and study the factors playing role in convergence, model quality and training efficiency. Further, we demonstrate how conditional computation can speed up the training \cite{bengio2015conditional} and how sparsely gating/routing each token through the network can efficiently be learned without any prior knowledge on task or language relatedness, exemplifying the capability of learning the routing decision directly from the data.

\subsection{Dataset and Baselines}

% \todo{Summary: a) realistic "in the wild" settings b) [15, 14] for dataset details.}
The premise of progressively larger models to attain greater quality necessitates large amounts of training data to begin with \cite{kaplan2020scaling}. Following the prior work on dense scaling for multilingual machine translation \cite{gpipe19,arivazhagan2019massively}, we committed to the realistic test bed of MT \textit{in the wild}, and use a web-scale in-house dataset. The training corpus, mined from the web \cite{10.5555/1873781.1873905}, contains parallel documents for 100 languages, to and from English, adding up to a total of 25 billion training examples.  A few characteristics of the training set is worth mentioning. Having mined from the web, the joint corpus is considerably noisy while covering a diverse set of domains and languages. Such large coverage comes with a heavy imbalance between languages in terms of the amount of examples per language pair. This imbalance follows a sharp power law, ranging from billions of examples for high-resourced languages to tens of thousands examples for low-resourced ones. While the above mentioned characteristics constitute a challenge for our study, it also makes the overall attempt as realistic as possible. We refer reader to \cite{gpipe19,arivazhagan2019massively} for the additional details of the dataset being used.

% \todo{Summary: a) baselines b) how we visualize quality gains.}
We focus on improving the translation quality (measured in terms of BLEU score \cite{papineni2002bleu}) from all 100 languages to English. This resulted in approximately 13 billion training examples to be used for model training\footnote{Compared to prior work using the same dataset, Kazakh and Latin to English language pairs were excluded from evaluation.}. In order to form our baselines, we trained separate bilingual Neural Machine Translation models for each language pair (e.g. a single model for German-to-English), tuned depending on the available training data per-language\footnote{We tuned batch-size and different values of regularization methods (e.g. dropout) in a Transformer-Big or Transformer-Base layout, for high or low-resourced languages respectively.}. Rather than displaying individual BLEU scores for each language pair, we follow the convention of placing the baselines along the $x$-axis at zero, and report the $\Delta$BLEU trendline of each massively multilingual model trained with \codename (see Figure~\ref{fig:data}). The $x$-axis in Figure~\ref{fig:data} is sorted from left-to-right in the decreasing order of amount of available training data, where the left-most side corresponds to high-resourced languages, and low-resourced languages on the right-most side respectively. To reiterate, our ultimate goal in universal machine translation is to amass the $\Delta$BLEU trendline of a single multilingual model above the baselines for all languages considered. We also include a variant of dense 96 layer Transformer Encoder-Decoder network T(96L) trained with GPipe pipeline parallelism on the same dataset as another baseline (dashed trendline in Figure~\ref{fig:data}). Training to convergence took over 6 weeks on 2048 \TPUv cores \footnote{T(96L) measured to be processing 1+ trillion tokens at 300k steps, processing around 4M tokens/step, total budget of 235.5 \TPUv core years}, outperforming the original GPipe T(128L)\footnote{64 encoder + 64 decoder layers, 16384 hidden dim, 32 attention heads}~\cite{gpipe19} and is the strongest single dense model baseline we use in our comparisons.

\subsection{Sparsely-Gated MoE Transformer: Model and Training}

Scaling Transformer architecture has been an exploratory research track recently \cite{Bapna_2018,Irie_2019,Wang_2019}. Without loss of generality, emerging approaches follow scaling Transformer by stacking more and more layers \cite{Bapna_2018,gpipe19}, widening the governing dimensions of the network (i.e. model dimension, hidden dimension or number of attention heads) \cite{devlin2018bert,raffel2019exploring} and more recently learning the wiring structure with architecture search \cite{so2019evolved}~\footnote{Since the approaches utilizing architecture search are compute intensive, they are not considered within the scope of this work.}. For massively multilingual machine translation, \cite{gpipe19} demonstrated the best practices of scaling using GPipe pipeline parallelism; in which a 128 layer Transformer model with 6 billion parameters is shown to be effective at improving high-resource languages while exhibiting the highest \textit{positive transfer} towards low-resource languages. Although very promising, and satisfying our desiderata for universal translation, dense scaling of Transformer architecture has practical limitations which we referred in Section~\ref{sec:intro} under \textit{training efficiency}. %After developing a pipeline parallelism, GPipe enabled effective scaling, but even then the 128 layer - 6 billion parameter Transformer model took around 6-7 weeks to train using 2048 \TPUv cores.

We aim for practical training time and seek for architectures that warrant training efficiency. Our strategy has three pillars; increase the depth of the network by stacking more layers similar to GPipe~\cite{gpipe19}, increase the width of the network by introducing multiple replicas of the feed-forward networks (experts) as described in Section~\ref{sec:model-moe} and make use of learned routing modules to (sparsely) assign tokens to experts as described in Section~\ref{sec:sparse}. With this three constituents, we obtain an easy to scale, efficient to train and highly expressive architecture, which we call Sparsely-Gated Mixture-of-Experts Transformer or MoE Transformer in short.

\paragraph{Model Details} To detail the model specifics, each expert is designed to have the same shape of a regular Transformer feed-forward network, and experts (MoE layers) are distributed once in every other Transformer layer. We tied the number of devices used for training to the number of experts per MoE layer for simplicity, although this is not a requirement. During training, we use float32 for both model weights and activations in order to ensure training stability. We ran additional scalability experiments with \MoE{2048}{60} with bfloat16~\cite{bfloat16} activations with total of 1 trillion model weights. Although trainable by careful and manual diagnostics, with deep 1 trillion model we encountered several trainability issues with numerical stability, hence did not include the results for the sake of reproducibility. For more model and training details, please see Appendix~\ref{sec:mt-more}.

% \paragraph{Adjusting the Batch Size}
% %\todo{need to rephrase to sound less inconsistent.}

% % {TPU engine attempts to strategically re-compute certain operators to fit the model in memory (called rematerialization} 
% Our original approach was to keep the $N/D$ number of tokens per device fixed to 2048 tokens, which is equivalent to fixing the number of processed tokens per expert in MoE layer. With batch size $N$ being 250k, 1 million and 4 million for 128, 512, 2048 experts respectively. This resulted in instability of training of 128 expert models, so we scaled up their batch size $N$ to 1 million.

% When training deeper 36L models activations and gradients consume more device memory than model weights\footnote{\MoE{512}{36} total per device memory usage is 13.3G, out of that 7.3G is activations and gradients.}, with little device memory left unused. So, there is a limited capability to increase per-device batch size further. 
% % Doubling the batch generally results in proportional increase of step time

\newcommand{\twolines}[2]{\begin{tabular}[c]{@{}c@{}}#1\\#2\end{tabular}}
% TODO: replace ones below
\newcommand{\expertsperlayer}{\begin{tabular}[c]{@{}c@{}}Experts/\\layer\end{tabular}}
\newcommand{\expertstotal}{\begin{tabular}[c]{@{}c@{}}Experts\\total\end{tabular}}
\newcommand{\totallayers}{\begin{tabular}[c]{@{}c@{}}Enc+Dec\\layers\end{tabular}}

% historic caption
% Test set quality scores of GShard - MoE Transformer models measured in $\Delta$BLEU, with a collection of 100 bilingual baselines (Transformer Big/Base) models. Bilingual models trained for each language pair are collapsed into the $x-axis$ at zero BLEU score and we report the improvements on top of the baselines. The quality of massively multilingualMoE Transformer models trained with GShard aremodels are reported with a separate (solid) trend-lines. in increasing order of number of parameters. Dashed trend-line represents a single 96 layer multilingual Transformer model T(96L) trained with GPipe on same dataset.For completeness, a massively multilingual GPipe model quality is also reported (dashed trend-line). Each trend-line is smoothed by a sliding window of 10 for clarityEach trend-line is smoothed by a moving average with a window size of 10 in order to better demonstrate quality difference for high, mid and low-resource languages.

\begin{figure}[t!]
\begin{center}
\includegraphics[scale=0.5]{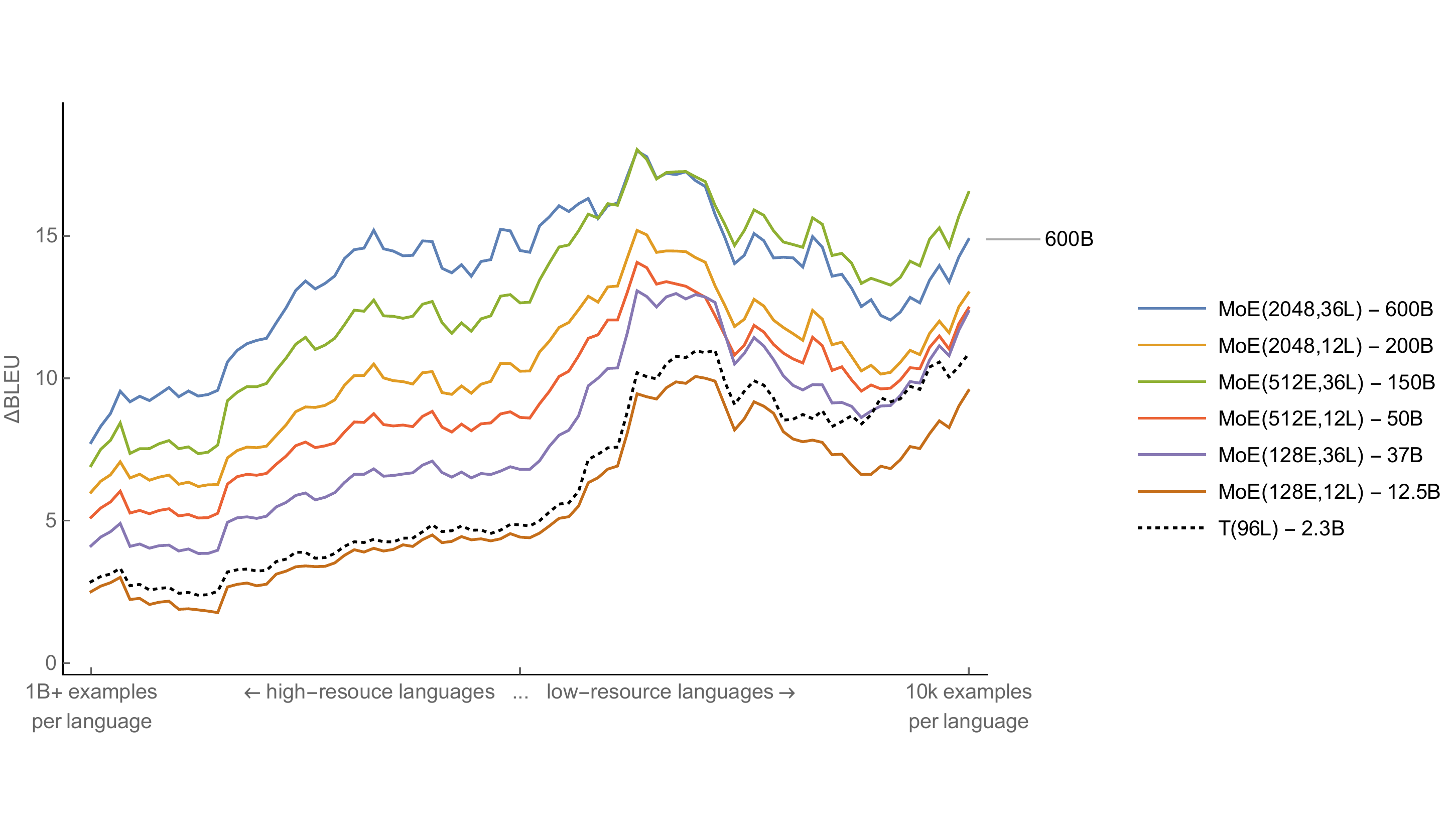}
\begin{tabular}{|c|l|r|r|r|}
\hline
Id & Model  &\twolines{BLEU}{avg.} &\twolines{$\Delta$BLEU}{avg.} & Weights \\\hline
(1) &\MoE{2048}{36}  & \textbf{44.3} & 13.5 & 600B \\
(2) &\MoE{2048}{12}  & 41.3 & 10.5 & 200B \\
(3) &\MoE{512}{36}   & 43.7 & 12.9 & 150B \\
(4) &\MoE{512}{12}   & 40.0 & 9.2 & 50B\\
(5) &\MoE{128}{36}   & 39.0 & 8.2 & 37B\\
(6) &\MoE{128}{12}   & 36.7 & 5.9 & 12.5B\\\hline
*   &T(96L)          & 36.9 & 6.1 & 2.3B\\
*   &Baselines       & 30.8 & - & 100$\times$0.4B\\\hline
\end{tabular}

\caption{Translation quality comparison of multilingual MoE Transformer models trained with GShard and monolingual baselines. Positions along the $x$-axis represent languages, raging from high- to low-resource. $\Delta$BLEU represents the quality gain of a single multilingual model compared to a monolingual Transformer model trained and tuned for a specific language. MoE Transformer models trained with GShard are reported with solid trend-lines. Dashed trend-line represents a single 96 layer multilingual Transformer model T(96L) trained with GPipe on same dataset. Each trend-line is smoothed by a sliding window of 10 for clarity. (Best seen in color)}
\label{fig:data}
\end{center}
\end{figure}

% BLEU	0.366788	0.390428	0.3999	0.436862	0.413405	0.442869	0.3080	0.3694														
% ΔBLEU	0.05881603921	0.08245603921	0.0919	0.1288900392	0.1054330392	0.1348970392	-	0.0614																						

% Describe a family of experiments:
% 	Comprehensive description of dataset, baselines and training settings including optimizer hyperparameters are presented in the Appendix.
% 	- Single model
% 			- additional FLOPS for dispatch / combine
% 				- no additional FLOPS to transform activations themselves
% 			- additional network  
% 		- layers
% 			- linear increase in FLOPS / network and generally step time

\subsection{Results}

Before going into the details of \textit{training efficiency}, we first investigate the effect of various design choices on building  MoE Transformer. In order to prune the search space, we explored varying two variables, number of layers in the Transformer encoder-decoder stack (L) and the total number of experts used for every other MoE layer (E). For depth, we tested three different options, 12 (original Transformer depth, which consists of 6 encoder and 6 decoder layers), 36 and 60 layers. For the number of experts that replaces every other feed-forward layer, we also tested three options, namely 128, 512 and 2048 experts. Note that, the number of devices used for training, is fixed to be equal to the number of experts per-layer, using 128, 512 and 2048 cores respectively independent of the depth being experimented. Please also see the detailed description in Table~\ref{tab:moe_family} for model configurations.

For each experiment (rows of the Table~\ref{tab:moe_family}), we trained the corresponding MoE Transformer model until it has seen 1 trillion ($10^{12}$) tokens. The model checkpoint at this point is used in the model evaluation. We did not observe any over-fitting patterns by this point in any experiment. Instead, we observed that the training loss continued to improve if we kept training longer. We evaluated BLEU scores that the models achieved for all language pairs on a held-out test set. Figure~\ref{fig:data} reports all our results.

Here we share a qualitative analysis for each experiment and discuss the implication of each setup on high-{} and low-resource languages in order to track our progress towards universal translation. To ground the forthcoming analysis, it is worth restating the expected behavior of the underlying quality gains. In order to improve the quality for both high-{} and low-resource languages simultaneously within a single model, scaled models must mitigate \textit{capacity bottleneck} issue by allocating enough capacity to high-resource tasks, while amplifying the \textit{positive transfer} towards low-resource tasks by facilitating sufficient parameter sharing. We loosely relate the expected learning dynamics of such systems with the long-standing memorization and generalization dilemma, which is recently studied along the lines of width vs depth scaling efforts \cite{Cheng_2016}. Not only do we expect our models to generalize better to the held-out test sets, we also expect them to exhibit high transfer capability across languages as another manifestation of generalization performance \cite{lampinen2018analytic}.

% What's the section structure?
%   
%   quantitative results ~ we talk about $\Delta$BLEU
%       performance depends on shape as well, e.g. 36L models have average size of the gated / activated subnetwork is 3x larger compared to corresponding shallower 12L models with same number of experts
%
%       size of activated subnetwork per sentennce is unknown, roughly proportional to sentence length in training due to extremely limited fractional expert capacity C=2 in 2048 layer model, it's ok to assume that all tokens are routed to different experts on each layer
%
%   sample efficiency is separate?
%   
%   

\begin{table}[t!]
\centering
\begin{tabular}{|c|l|r|r|r|r|r|r|}
\hline
Id & Model & \twolines{Experts}{Per-layer} & \expertstotal & \twolines{\TPUv}{Cores} & \totallayers & Weights \\\hline \hline
(1)&\MoE{2048}{36}  & 2048  & 36684 & 2048    & 36 & 600B \\
(2)&\MoE{2048}{12}  & 2048  & 12228 & 2048    & 12 & 200B \\
(3)&\MoE{512}{36}   & 512   & 9216  & 512     & 36 & 150B \\
(4)&\MoE{512}{12}   & 512   & 3072  & 512     & 12 & 50B \\
(5)&\MoE{128}{36}   & 128   & 2304  & 128     & 36 & 37B \\
(6)&\MoE{128}{12}   & 128   & 768   & 128     & 12 & 12.5B \\\hline
*  &\MoE{2048}{60}  & 2048  & 61440 & 2048    & 60 & 1T \\\hline
\end{tabular}
\vspace{0.1in}
\caption{MoE Transformer model family. To achieve desired capacity we \textit{i)} increased the depth by stacking more layers, \textit{ii)} increased the width of the network by scaling the number of experts per MoE layer along with number of cores used for training.}
\label{tab:moe_family}
\end{table}

\paragraph{Deeper Models Bring Consistent Quality Gains Across the Board}

We first investigate the relationship between the model depth and the model quality for both high- and low-resource languages. Three different experiments are conducted in order to test the generalization performance, while keeping the number of experts per-layer fixed. With an increasing number of per-layer experts for each experiment (128, 512 and 2048), we tripled the depth of the network for each expert size, from 12 to 36. This resulted in three groups where experts per-layer fixed but three times the depth within each group:

% \begin{enumerate}
% \item 128 experts per-layer (E) with 12 vs 36 layers (L), brown vs purple trend-lines in Figure~\ref{fig:data},
% \item 512E with 12L vs 36L, red vs green trend-lines,
% \item 2048E with 12L vs 36L, orange vs blue trend-lines.
% \end{enumerate}

\noindent For each configuration shown in Fig.~\ref{fig:data}, we observed that increasing the depth (L) while keeping the experts per-layer (E) fixed, brings consistent gains for both low and high resourced languages (upwards $\Delta$ shift along the $y$-axis), almost with a constant additive factor every time we scale the depth from 12L to 36L  (2-to-3 BLEU points on average as shown in the last column of Table~\ref{tab:m4_results}). %It should be noted that, compared to 12L models, 36L models have approximately 3-times larger activated sub-networks for each token and consequently for each sentence, which is expected to be one of the contributors of the observed quality boost.

\paragraph{Relaxing the Capacity Bottleneck Grants Pronounced Quality Gains}
\label{sec:bottleneck}

Earlier in Section~\ref{sec:mnmt} we highlighted the influence of the \textit{capacity bottleneck} on task interference, resulting in degraded quality especially for high resourced languages. Later we alleviated this complication by increasing the number of experts per-layer, which in return resulted in a dramatic increase in the number of parameters (weight) of the models studied. Here we investigate whether this so called \textit{capacity bottleneck} is distinctly observable and explore the impact on model quality and efficiency once it is relaxed. To that end, we first consider three models with identical depths (12L), with increasing number of experts per-layer: 128, 512 and 2048. As we increase the number of experts per-layer from 128 to 512 by a factor of four, we notice a large jump in model quality, +3.3 average BLEU score across 100 languages. However again by four folds scaling of the number of experts per-layer, from 512 to 2048, yields only +1.3 average BLEU scores. Despite the significant quality improvement, this drop in gains hints the emergence of diminishing returns. 

Speculatively, the capacity bottleneck is expected to be residing between 128 to 512 experts, for the particular parametrization, number of languages and the amount of training data used in our experimental setup. Once the bottleneck is relaxed, models enjoy successive scaling of the depth, which can be seen by comparing 12 versus 36 layer models both with 128 experts. Interestingly increasing the depth does not help as much if the capacity bottleneck is not relaxed.

%Notably, $\Delta$BLEU gains for \MoE{512}{36} exceed ones with higher capacity, but shallower \MoE{2048}{12}. While a comparison of proportionally smaller models, shows that \MoE{128}{36} is suboptimal compared to \MoE{512}{12}. One can conclude that scaling depth brings most quality gains only after capacity bottleneck is resolved. 

% talk about sample efficiency of 36L models?
% talk about 3x training / inference cost increase? 

\paragraph{Having More Experts Improve Quality Especially for High-Resourced Tasks} 

Another dimension that could shed light on the quality gains of scaling in multi-task models is the contrast between high and low resource language improvements. As mentioned before, low resourced languages benefit from transfer while high resource languages seek for added capacity. Next we examine the effect of increasing the experts per-layer while fixing the depth. 

%We analyze two scenarios, first fixing the layer number to 12 and increasing the number of experts-per layer, and second repeating it with 36 layers. 
As can be seen in Figure~\ref{fig:data}, for 12 layer models increase in the expert number yields larger gains for high resourced languages as opposed to earlier revealed diminishing returns for low-resourced languages. A similar pattern is observed also for 36 layer models. %but with a more noticeable trade-off between high and low-resource languages. 
While adding more experts relaxes the capacity bottleneck, at the same time it reduces the amount of transfer due to a reduction of the shared sub-networks.
%as the gains for high resource languages increase, the gains for low resourced languages diminish. Unsurprisingly, such behavior is expected with the hypothesis that, 

\paragraph{Deep-Dense Models are Better at Positive Transfer towards Low-Resource Tasks}

Lastly we look into the impact of the depth on low-resourced tasks as a loose corollary to our previous experiment. In order to do so, we include a dense model with 96 layers T(96L) trained with GPipe on the same data into our analysis. % using larger number of chips for training. 
We compare T(96L) with the shallow \MoE{128}{12} model. While the gap between the two models measured to be almost constant for the majority of the high-to-mid resourced languages, the gap grows in favor of the dense-deep T(96L) model as we get into the low-resourced regime. Following our previous statement, as the proportion of the shared sub-networks across tasks increase, which is 100\% for dense T(96L), the bandwidth for transfer gets maximized and results in a comparably better quality against its shallow counterpart. Also notice that, the same transfer quality to the low-resourced languages can be achieved with \MoE{36}{128} which contains 37 billion parameters.

We conjecture that, increasing the depth might potentially increase the extent of transfer to low-resource tasks hence generalize better along that axis. But we also want to highlight that the models in comparison have a disproportionate training resource requirements. We again want to promote the importance of \textit{training efficiency}, which is the very topic we studied next.
%: while T(96L) model occupies a whole TPU-pod for 6 weeks (2048 cores), the GShard models take less than 17 days, using significantly less amounts of resources (128 cores). We again want to promote the importance of \textit{training efficiency}, which is the very topic we studied next.

%\begin{enumerate}
    %\item Improve on top-10(top-5?) high resource languages compared to deep models, previously there was not enough capacity for such tasks with high resource.
    %\item Deep models demonstrated downstream transfer to low- and med- resource languages but high resource tasks suffered from interference due to limited capacity(it'a an unsupported hypothesis?)
    %\item GPipe 128 added on average X BLEU for top10 high resource comparedto Y BLEU for 2048-36L    
%\end{enumerate}
%%%

\subsection{Training Efficiency}

In this section we focus on the \textit{training efficiency} of MoE Transformer models. So far, we have seen empirical evidence how scaling the models along various axes bring dramatic quality gains, and studied the factors affecting the extent of the improvements. In order to measure the \textit{training efficiency}, we first keep track of the number of tokens being processed to reach a certain training loss and second we keep track of the wall-clock time for a model to process certain number of tokens.
Note that, we focus on the training time and training loss\footnote{Training loss reported in this section corresponds to cross-entropy loss and excludes the auxiliary loss term introduced in Section~\ref{sec:model-moe}} while varying other factors, as opposed to test error, which we analyzed in the previous section.
% We want to also remind that, early stopping is not a feasible option for the models under investigation since the stopping decision is made by the number examples being processed by each model. 

%The largest two contributors to the cost of building large scale models are i. the number of compute devices being utilized, measured in terms of number of cores being used and ii. the duration that these devices are being utilized, measured in terms of the wall-clock time spent to reach a certain quality level. In this section we focus on the latter item, and first time how many steps it takes different models to reach varying training loss thresholds, and second compare the training loss behavior of each model, while controlling the amounts of training data being ingested. While doing so, we want to measure both sample- and wall-time efficiency of models which we found to be two important variables effecting \textit{training efficiency} of models at scale. 

\paragraph{Deeper models are more sample efficient, converge faster with fewer examples}
It has been shown that, deeper models are better at sample efficiency, reaching better training/test error given the same amount of training examples \cite{gpipe19,shoeybi2019megatron}, commonly attributed to the acceleration effect of over-parametrization \cite{arora2018optimization}. We empirically test the hypothesis again using GShard with MoE Transformers and share trade-offs for models that are not only deep, but also sparsely activated. 

For this purpose, we compare number of tokens being processed by each model to reach a preset training loss. A general trend we observe from Table~\ref{tab:ce} is that, MoE Transformer models with 3 times the depth need 2 to 3 times fewer tokens to reach the preset training loss thresholds. For example \MoE{128}{12} takes 3 times the number of tokens to reach 0.7 training cross-entropy compared to \MoE{128}{36}, (6) vs (5). We observe a similar trend for models with 512 and 2048 experts, (4) vs (3) and (2) vs (1).

%The similar tre comparing (6)-(5), (4)-(3) and (2)-(1) from Table~\ref{tab:ce}. Note that, each comparison is across models using same number of TPU cores with identical batch sizes. 

\begin{table}[ht]
\centering
\begin{tabular}{|c|l|r|r|r|r|}
\hline
\multirow{2}{*}{Id} & \multirow{2}{*}{Model} & \multirow{2}{*}{Cores} & \multicolumn{3}{c|}{\twolines{Billion tokens to}{cross-entropy of}} \\ \cline{4-6} 
    &                &       & 0.7  & 0.6   & 0.5   \\ \hline\hline
(1) &\MoE{2048}{36}  &  2048 &82   & 175  & 542 \\
(2) &\MoE{2048}{12}  &  2048 &176  & 484  & 1780 \\
(3) &\MoE{512}{36}   &  512  &66   & 170  & 567  \\
(4) &\MoE{512}{12}   &  512  &141  & 486  & -     \\
(5) &\MoE{128}{36}   &  128  &321  & 1074 & -     \\
(6) &\MoE{128}{12}   &  128  &995  & -     & -     \\
\hline
\end{tabular}
\caption{The number of tokens have been seen by a model during training to reach three different cross-entropy loss. A general trend is that deeper models are more sample efficient and converge faster than the comparable shallow ones.}
\label{tab:ce}
\end{table}

Another intriguing observation from Table~\ref{tab:ce}, is again related to the presence of \textit{capacity bottleneck}. Comparing the models with same depth, (5), (3) and (1), we notice a significant drop in the number of tokens required to reach training loss of 0.7, as we transition from 128 to 512 number of experts. Practically that is where we observed the capacity bottleneck was residing, aligning with the hypothesis in Section \ref{sec:bottleneck}. After this phase shift, models with ample capacity tend to exhibit similar sample efficiency characteristics, as in models (3) and (1).

\paragraph{Largest model (600B) can be trained under 4 days achieving the best quality} Next we delve deeper into the interaction between model size and wall-clock time spent for training. We monitor number of TPU cores being used, training steps per-second, total number of tokens per batch, TPU core years\footnote{TPU core years is simply measured by the product of number of cores and wall-clock time in years.}, and actual wall-clock time spent in days for training (see Table~\ref{tab:m4_results} columns respectively). 

We start with investigating one of the largest models we trained, \MoE{2048}{36} with 600 billion parameters, model with id (1). Having utilized 2048 TPU cores for 4 days, this model achieves the best translation quality in terms of average BLEU, but also takes a total of 22.4 TPU years to train. While we have not seen any signs that the quality improvements plateau as we scale up our models, we strive for finding cost-effective solutions for scaling. 

Results in Table~\ref{tab:m4_results} again validates scaling with conditional computation is way more practical compared to dense scaling. Given the same number of TPU cores used by (1), the dense scaling variant, T(96L), appears to be taking more than ten times to train (235 TPU core years), while trailing behind in terms of model quality compared to models trained with GShard. 

%Following our aspiration for \textit{training efficiency}, we seek for a range of optimal configurations that minimizes TPU core years, training time in days, and total number of cores being utilized, while maximizing the final model quality. Given the same number of TPU cores used by (1), its 3-times shallower counterpart (2) comes out as a performant alternative but with a significant drop in model quality (BLEU column in Table~\ref{tab:m4_results}).

% BLEUs
% 0.366788	0.390428	0.3999	0.436862	0.413405	0.442869	0.3080
\begin{table}[h!]
\centering
\begin{tabular}{|c|l|r|r|r|r|r|r|}
\hline
Id & Model & Cores & \twolines{Steps}{per sec.} & \twolines{Batch sz.}{(Tokens)} & \twolines{TPU core}{years} & \twolines{Training}{time (days)} &\twolines{BLEU}{avg.} \\\hline \hline
(1) &\MoE{2048}{36}& 2048 & 0.72 & 4M & 22.4 & \textbf{4.0} & \textbf{44.3} \\
(2) &\MoE{2048}{12}& 2048 & 2.15 & 4M & 7.5 & 1.4 & 41.3 \\
(3) &\MoE{512}{36} & 512 & 1.05 & 1M & 15.5 & 11.0 & 43.7 \\
(4) &\MoE{512}{12} & 512 & 3.28 & 1M & 4.9 & 3.5  & 40.0 \\
(5) &\MoE{128}{36} & 128 & 0.67 & 1M & 6.1 & 17.3 & 39.0 \\
(6) &\MoE{128}{12} & 128 & 2.16 & 1M & 1.9 & 5.4 & 36.7 \\\hline
*   &T(96L) & 2048 & - & 4M & $\sim$235.5 & $\sim$42 & 36.9 \\\hline
\end{tabular}
\caption{Performance of MoE models with different number of experts and layers. }
\label{tab:m4_results}
\end{table}

In this section, we benchmarked GShard with MoE Transformers applications to multilingual machine translation (in particular to M4). We identified variables that are affecting the end result, such as \textit{capacity bottleneck}, \textit{positive transfer} and \textit{training efficiency}, and provided experimental results in order to reveal the interplay between them. Next we will delve deep into performance related topics of \codename, such as memory and runtime efficiency and communication benchmarks.

\section{Performance and Memory Consumption}
\label{sec:performance}

This section discusses how well \codename achieves computation and memory efficiency on the TPU platform. Our measurement and analysis show that the device memory consumption is roughly constant when we increase the number of devices and experts, and the step time grows sublinearly, i.e., 1.7x execution time increase when we scale the model by 16x from 128 devices to 2048 devices. We also provide microbenchmarks and analyses for a variety of partitioned operators, which could guide use cases beyond this paper.

% \subsection{Roofline Analysis}

% For each layer, Table \ref{tab:transformer_roofline} summarizes the total amount of memory, computation and communication requirement.

% \begin{table}[ht]
% \centering
% \begin{tabular}{|c|c|c|c|}
% \hline
%     & FLOPS  & Memory  & Network \\ \hline
% QKV Projection        & $B \times T \times M^2 \times 6$  & $B \times T \times M \times 3$       & 0 \\
% Output Projection     & $B \times T \times M^2 \times 2$  & $B \times T \times M$     & 0 \\
% Attention             & $B \times T^2 \times M\times 4$   & $B \times T \times (L + M)$ & 0 \\
% FeedForward Layers    & $B \times T \times M \times H \times 4$      & $B \times T \times (M + H)$    & 0 \\
% MoE Dispatch/Combine  & $B \times T \times M \times E \times 2$       & $B \times T \times E $   & $B \times T \times M \times 2$ \\ \hline
% \end{tabular}
% \vspace{0.1in}
% \caption{Roofline performance for MoE and non-MoE layers. For simplicity, we set the number of attention heads to 1, and attention hidden dimension to the same size as the model dimension $M$.}
% \label{tab:transformer_roofline}
% \end{table}

\subsection{Memory Efficiency and Scalability}
In the \codename model, there are mainly three types of memory usage, all of which have constant per-device sizes after SPMD partitioning, when the number of experts increases.

\begin{itemize}
    \item Replicated weights (e.g. transformer feed-forward layers).
    \item Distributed weights (MoE feed-forward layers\footnote{Gate projection weights are $O(E)$ in size and could be partitioned, but in practice they are small enough to be replicated and only have negligible effect on peak memory usage.}).
    \item Activations (output of each layer that is used in both forward and backward pass).
    % \item Programs (binary code that is used to execute on device), which is small compared to other types of memory.
\end{itemize}

\begin{figure}[t!]
\centering
\includegraphics[width=0.95\textwidth]{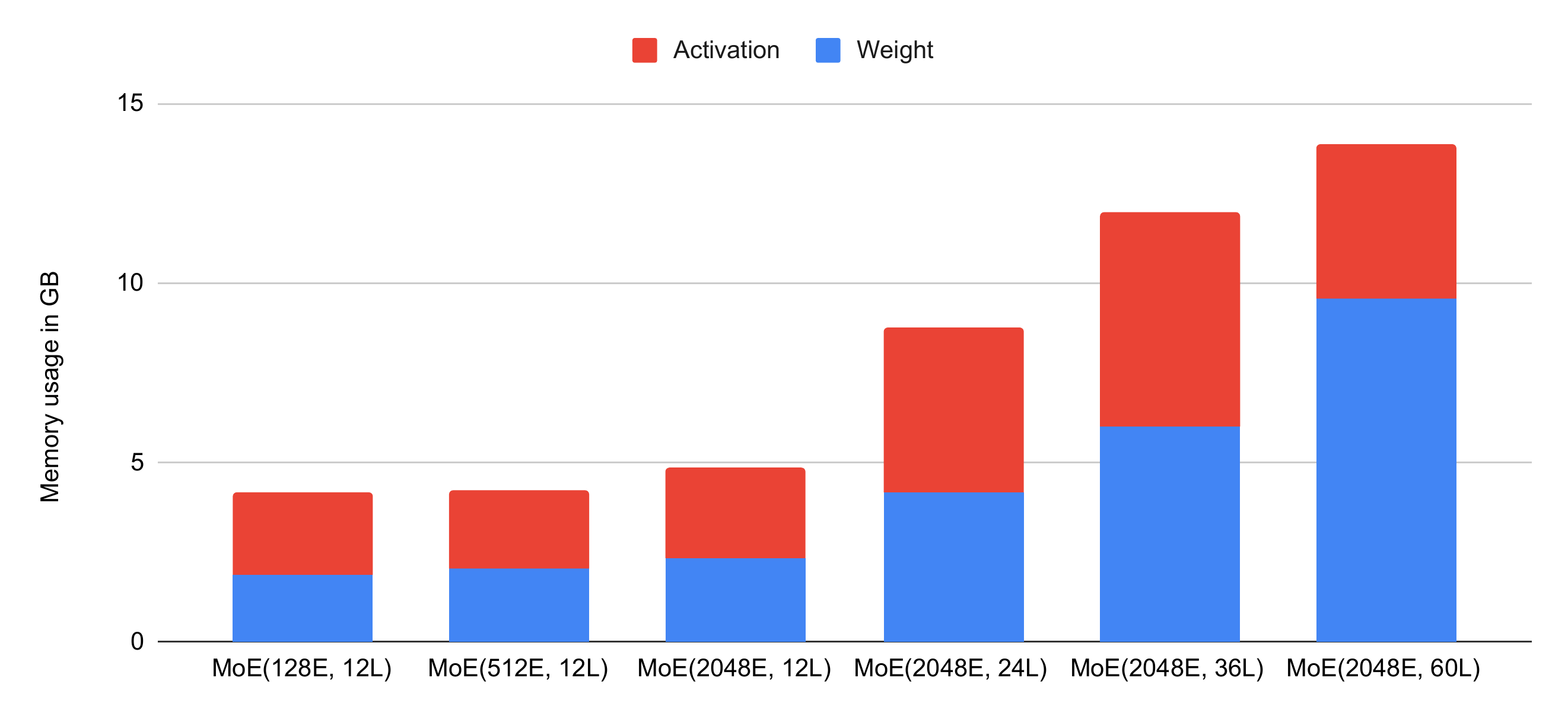}
\caption{Per-device memory consumption in gigabytes.}
\label{fig:perf-memory}
\end{figure}

The $O(1)$ memory scaling is demonstrated in Figure~\ref{fig:perf-memory}, which shows the per-device memory usage distribution for different models. With a fixed number of layers, both weight memory and activation memory stay constant when the number of experts increases.

On this other hand, weight memory and activation memory both scale linearly with the number of layers. When the memory requirement exceeds available memory on each device, compiler-based rematerialization will automatically recompute part of the activations in the backward pass in order to reduce peak activation memory. This is why the activation size for MoE(2048E, 60L) is smaller than MoE(2048E, 36L). The overhead of rematerialization is also optimized, e.g. only 28\% and 34\% of the total cycles are spent on recomputation for 36L and 60L models respectively, and 0\% for 12L and 24L since they fit in device memory without rematerialization.

%(yuanzx: this is not correct) For the replicated weights, another XLA optimization, weight-update sharding~\cite{xu2020automatic}, is automatically triggered to only maintain a shard for each replication to further reduce the weight requirement.

\subsection{Runtime Efficiency and Scalability}\label{sec:perf-scalability}

% \begin{figure}[htb]
% \includegraphics[width=0.7\textwidth]{perf_breakdowns.png}
% \centering
% \caption{Device time breakdown of gate computation}
% \label{fig:breakdown}
% \end{figure}

\begin{figure}[t!]
\centering
\includegraphics[width=0.95\textwidth]{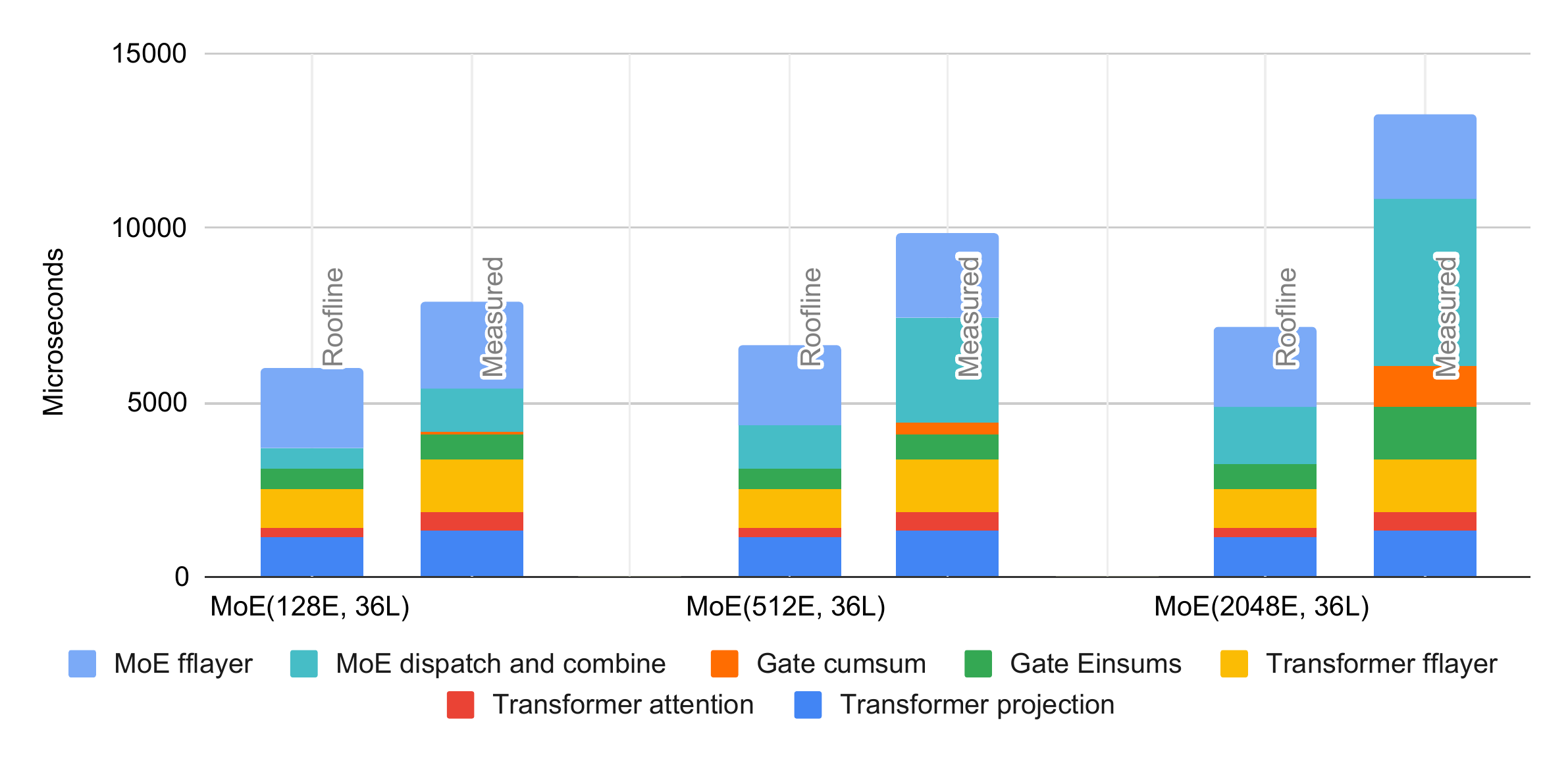}
\caption{Measured vs roofline execution time breakdown. Only the forward pass is shown, and the backward pass has similar breakdown. ``MoE dispatch and combine'' represents cross-partition communication with \texttt{AllToAll}.}\label{fig:perf-breakdown-roofline}
\end{figure}

Figure~\ref{fig:perf-breakdown-roofline} shows the breakdown of execution time for an MoE layer and its adjacent Transformer layer. It also compares the achieved performance to a roofline, which is estimated by assuming compute-, memory-, or communication-bounded operations can achieve 100\% of the peak FLOPS, memory bandwidth, or interconnect bandwidth. This is a very optimistic estimate as many operators are bounded by a mixed set of resources. At a smaller scale (128 experts), our model can achieve > 70\% of the roofline performance. The device time increases by 1.7x when we scale the model to 16x larger (2048 experts), and can still achieve 48\% of the roofline performance.

% Our MoE Transformer model generally has \texttt{[B, T, M]} layer activation tensors where $N=B*T$ is the number of tokens, \texttt{B} and \texttt{T} are batch and time dimensions. MoE layer internally re-groups token representations from \texttt{[B, T, M]} to \texttt{[G, S, M]} tensor and back.

% To dispatch and combine MoE layer activations of size $N \cdot M$ we multiply them by $G \cdot S \cdot E \cdot C$ mask and weight matrices, where $C = 2 \cdot S / E$ is fractional expert capacity, $G$ is the number of groups of size $S = N /G$, and $M$ is the model dimension. Computational cost of these operations is $O(N^2 / G)$ and is inversely proportional to $G$. For example, with 1M tokens per batch, 128 devices, 1 expert/device, 1 group/device it results in \textasciitilde68 GB\footnote{128 * 8192 * 128 * 128 * 4bytes} matrices and costly einsum operations. Disproportionate increase of $G$ leads to sub-optimal gating, so we select it keeping group size $S$ between 1024 and 2048 tokens.

Before analyzing performance scalability, we recall the size scaling of relevant tensor dimensions as discussed in Section~\ref{sec:moela}. With $D$ devices, the number of experts $E$ and the group count $G$ are both set to $O(D)$. The fractional per-group expert capacity $C$ is set to  $O(1/D)$. This setup cannot scale indefinitely, since $C$ needs to be at least 1, but it is good enough to scale to thousands of experts.

\paragraph{Transformer layers and MoE feed-forward layer} These are the dense parts of the model, which are designed to achieve peak TPU utilization. On each device, these computations also have a constant cost when we scale to more experts. Feed-forward layers and Transformer projections are mainly large matrix multiplications that utilize the TPU's matrix unit well. These operations have achieved > 85\% peak FLOPS in our experiment. The attention operations are composed of mainly batch matmuls, which are bounded by memory bandwidth when sequence lengths are small. As a result, in our experiments attention operations only achieved > 30\% peak FLOPS.

\paragraph{Gate computation} In Figure~\ref{fig:perf-breakdown-roofline}, ``Gate Einsum'' represents the first two and the last \texttt{Einsums} in Algorithm~\ref{alg:moe}. The first \texttt{Einsum} is the projection that calculates per-expert input to \texttt{softmax}. It has an $O(D)$ cost, but it is a very small part of the layer. The other two \texttt{Einsums} are dispatching tokens and combining expert results. They effectively implement \texttt{Gather} with one-hot matrices, which are more expensive, but with constant $O(GC)=O(1)$ cost that is independent from the number of experts. The execution time of these \texttt{Einsums} increases by around 2x when we scale from 128 to 2048 experts (16x).

The remaining per-device gating computation involves many general-purpose computations like \texttt{ArgMax} and \texttt{Cumsum}, which are either memory-bound or even sequential in nature, thus not designed to utilize TPUs well. The majority of the time is spent on sequential \texttt{Cumsum} operations to invert one-hot matrices that represent selected experts for each token to one-hot matrices that represent selected tokens for each expert. The linear complexity of \texttt{Cumsum} is demonstrated in Figure~\ref{fig:perf-breakdown-roofline}. This part of the gating computation also has an $O(D)$ cost, but fortunately, similar to the \texttt{Einsum} before \texttt{softmax}, it has a very small constant factor. It has negligible execution time with 128 experts, and takes less than 10\% of the total time spent in the MoE and Transformer layers with 2048 experts.

The most significant part of gating is communication, shown as ``MoE dispatch and combine'' in Figure~\ref{fig:perf-breakdown-roofline}. These are \texttt{AllToAll} operators, and as we will discuss in Section~\ref{sec:microbench}, their cost is $O(\sqrt{D})$. When the number experts grows 16x from 128 to 2048, the execution time increases by about 3.75x, and their proportion of execution time in the MoE and Transformer increases from 16\% to 36\%.

\subsection{Communication Microbenchmarks and Per-Operator Scalability}\label{sec:microbench}
In this section, we measure and analyze the performance scalability of the SPMD partitioner for basic operators, which can be used to guide use cases beyond the MoE model presented in this paper.

\begin{figure}[t!]
\begin{center}
\includegraphics[width=0.9\textwidth]{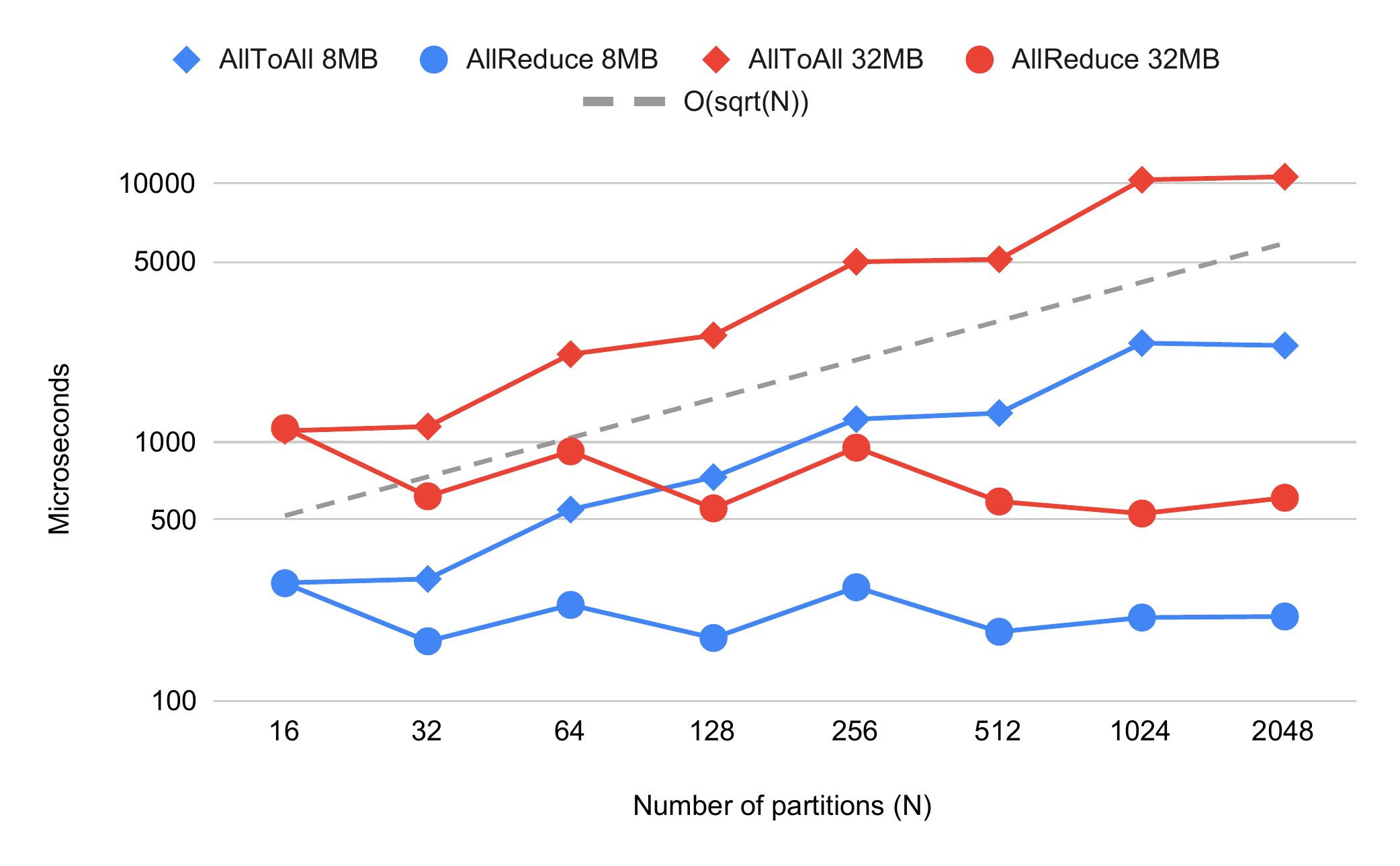}
\caption{Performance scaling of communication, \texttt{AllReduce} and \texttt{AllToAll}. Log scale on both axes. \texttt{AllReduce} cost is roughly $O(1)$, and \texttt{AllToAll} cost is roughly $O(\sqrt{D})$, where $D$ is the number of partitions. We measure their performance with 8MB and 32MB data. For \texttt{AllToAll}, that means each partition initially has 8MB (or 32MB) data, then divides it to $D$ pieces, and sends each piece to a different receiving partition.}
\label{fig:collective_microbench}
\end{center}
\end{figure}

\paragraph{Performance scaling of communication primitives} Two critical collective communication operators in the MoE model are \texttt{AllReduce} and \texttt{AllToAll}. \texttt{AllReduce} is used in accumulating partial results, and \texttt{AllToAll} is used in resharding (Section~\ref{sec:op-partitioner}). Figure~\ref{fig:collective_microbench} shows their performance scalability from 16 to 2048 partitions. \texttt{AllReduce} on TPU has an execution time independent from the number of devices. The variance in Figure~\ref{fig:collective_microbench} is due to specifics of each topology, e.g., whether it is a square or a rectangle, and whether it is a torus or a mesh.

\texttt{AllToAll}, on the other hand, gets more expensive as the number of partitions grows, but in a sublinear manner. On our 2D TPU cluster, \texttt{AllToAll} cost is roughly $O(\sqrt{D})$, where $D$ is the number of partitions. This is because with a fixed amount of data each partition sends (8MB or 32MB in Figure~\ref{fig:collective_microbench}), the total amount of data that all partitions send is $d=O(D)$. Meanwhile, each data piece needs to travel $h=O(\sqrt{D})$ hops on average, and there are overall $l=O(D)$ device-to-device links in the network. Therefore, if it is bandwidth-bound, the execution time of an \texttt{AllToAll} is
\[
t=\frac{dh}{l}=O(\frac{D\sqrt{D}}{D})=O(\sqrt{D}).
\]
Even if it is latency-bound, the execution time will still be $O(h)=O(\sqrt{D})$.
Comparing 2048 partitions and 16 partitions, while $D$ grows by 128 times, the execution time of \texttt{AllToAll} only increases by 9 times. This enables us to use resharding to efficiently implement cross-partition dispatching (Figure~\ref{fig:einsum_example}).

\texttt{AllGather} and \texttt{CollectivePermute} are easier to analyze.
\texttt{AllGather}'s output is $D$ larger than the input, and if we fix input size, then its communication cost is $O(D)$. \texttt{CollectivePermute} has a one-to-one communication pattern, and with reasonable device arrangement where the source-destination pairs are close, its cost is $O(1)$ for a fixed input size.

\begin{table}[ht]
\centering
\begin{tabular}{|c|c|c|c|r|}
\hline
    & $O(D)$ & Total & \multicolumn{2}{c|}{Per-partition}  \\\cline{4-5}
    & Dimensions & Compute & Compute  & Communication \\ \hline
\texttt{Add(\Sharded{A},\Sharded{A}->\Sharded{A})} & \texttt{A} & $O(D)$ & $O(1)$ & 0 \\ \hline
\texttt{Matmul(A\Sharded{B},\Sharded{B}C->AC)} & \texttt{B} & $O(D)$ & $O(1)$ & $O(1)$ \texttt{AR} \\ \hline
\texttt{Matmul(\Sharded{A}B,BC->\Sharded{A}C)} & \texttt{A} & $O(D)$ & $O(1)$ & 0 \\ \hline
\texttt{Matmul(\Sharded{A}B,\Sharded{B}C->\Sharded{A}C)} & \texttt{A,B} & $O(D^2)$ & $O(D)$ & $O(D)$ \texttt{AG} or \texttt{CP}\\ \hline
\texttt{Matmul(\Sharded{A}B,B\Sharded{C}->A\Sharded{C})} & \texttt{A,C} & $O(D^2)$ & $O(D)$ & $O(D)$ \texttt{AG} or \texttt{CP}\\ \hline
\texttt{Reduce(\Sharded{A}B->\Sharded{A})} & \texttt{A} & $O(D)$ & $O(1)$ & 0 \\ \hline
\texttt{Reduce(\Sharded{A}B->B)} & \texttt{A} & $O(D)$ & $O(1)$ & $O(1)$ \texttt{AR} \\ \hline
\texttt{Einsum(\Sharded{G}SEC,\Sharded{G}SM->\Sharded{E}GCM)} & \texttt{G,E} * & $O(D)$ & $O(1)$ & $O(\sqrt{D})$ \texttt{AA} \\ \hline
\texttt{Convolution(BI\Sharded{X}Y,xyIO->BO\Sharded{X}Y)} & \texttt{X} ** & $O(D)$ & $O(1)$ & $O(1)$ \texttt{CP} \\ \hline
\end{tabular}
\vspace{0.1in}
\caption{Scalability of partitioned operators. Abbreviation for communication primitives: \texttt{AR: AllReduce}, \texttt{AG: AllGather}, \texttt{CP: CollectivePermute}, \texttt{AA: AllToAll}. *This is the dispatch \texttt{Einsum} in our model, where we set \texttt{C} to $O(1/D)$. **\texttt{I/O} are the input/output feature dimensions, \texttt{B} is the batch dimension, \texttt{X/Y} are input spatial dimensions, and \texttt{x/y} are the kernal spatial dimensions.}
\label{tab:op-scalability}
\end{table}

%\todo{yuanzx hyouklee, more explanation}
\paragraph{Partitioned operator scalability} We summarize the performance scalability for common operators using \codename in Table~\ref{tab:op-scalability}. It contains the \texttt{Einsum}/\texttt{Matmul} examples in Section~\ref{sec:op-partitioner}, and also other common operators like \texttt{Convolution} and \texttt{Reduce}. The table includes the local compute on each partition, as well as the required communication based on our analysis above.

Most operators in Table~\ref{tab:op-scalability} have sublinear scalability in terms of both compute and communication, which is consistent with our performance measurement of the MoE model. The $O(1)$ scaling of spatially partitioned convolutions also demonstrates the efficiency of \codename for image partitioning (Appendix~\ref{sec:appendix-spmd-conv}).

However, the last two \texttt{Matmul} operators in Table~\ref{tab:op-scalability} have $O(D)$ scaling of per-partition compute and communication, where they have unmatched sharding in the operands. This is not due to inefficiency in the partitioning algorithm, but because the total compute in the full operator is very large ($O(D^2)$). Different partitioning strategies can be used for these cases, producing different communication primitives: replicating one operand will result in \texttt{AllGather} (requiring the replicated operand to fit in device memory), while slicing in a loop (Figure~\ref{fig:cannon_example}) will result in \texttt{CollectivePermute}.

% \subsection{Host-Device Setup for Variables and Input}
% So far we have focused on the performance of TPU device computations, and now we briefly discuss our host-device setup that avoids bottlenecks on the host.

% \paragraph{Weights distribution and checkpointing} The TPU programming model that is used by this work keeps weights on the host. Weights are sharded and distributed to devices every time a session is invoked. This makes checkpointing convenient as the master copy of weights always lives on the host. However, when scaling the model to more than 100B, host memory may not be large enough to hold all the variables. As a result, we store variables in a distributed manner, i.e. for each layer, the variable only has one master copy and lives on a different host. This way the host memory requirement is distributed and checkpointing can be done in parallel.

% \paragraph{Input pipeline parallelization} Normally for translation models, the input size is small thus can be processed serially in a single machine. However, when scaling to >2048 TPU cores, the processing power is so strong that the host becomes bottleneck even for translation task. As a result, we parallelize the input pipeline to multiple machines to avoid host bottleneck.

\section{Related Work}~\label{sec:related}

\textbf{Neural networks} Deep learning models have been very successful in advancing sub-fields
of artificial intelligence. For years, the fields have been continuously reporting new state of 
the art results using varieties of model architectures for computer vision tasks~\cite{krizhevsky2012imagenet,szegedy2015going,he2016deep}, 
for natural language understanding tasks~\cite{sutskever2014sequence,bahdanau2014neural,wu2016google},
for speech recognition and synthesis tasks~\cite{hinton2012deep,chan2016listen,chiu2018state,oord2016wavenet,shen2018natural}.
More recently, attention-based Transformer models further advanced state of the art of these fields~\cite{vaswani2017attention,devlin2018bert}.

\textbf{Model scaling} Both academic research and industry applications observed that larger neural networks tend to perform better on large enough datasets and for complex tasks. Within a single model family, simply making the network wider or deeper often improves the model quality empirically. E.g., deeper ResNets performed better~\cite{he2016identity}, bigger Transformer models achieved better translation quality~\cite{vaswani2017attention}, models with larger vocabulary, or embedding or feature crosses work better, too~\cite{arivazhagan2019massively,conneau2019unsupervised}.  Across different model families, it has also been observed that bigger models with larger model capacities not only fit the training data better but also generalize better on test time~\cite{45820,neyshabur2017exploring,gpipe19}. 
This observation motivated many research efforts to build much bigger neural networks than those typically used in deep learning research models or production models. Shazeer et al. showed that a recurrent language model with 69 billion parameters using mixture-of-expert layers achieved much lower test perplexity for the one billion words (LM1B) benchmark~\cite{shazeer2017outrageously}. Brown et al. showed that a non-sparse 175 billion parameters model is capable of exhibiting highly accurate few-shot performance on several downstream NLP tasks.

\textbf{Hardware} Neural networks demand non-negligible amounts of computation power. To address such a demand, special hardware (chips and networked machines) built for neural network training and inference can be dated back to 25 years ago~\cite{ienne1996special}. Since late 2000s, researchers started to leverage GPUs to accelerate neural nets~\cite{raina2009large,krizhevsky2012imagenet,cirecsan2010deep}. More recently, the industry also invested heavily in building more dedicated hardware systems chasing for more cost-effective neural network hardware~\cite{jouppi2017datacenter}. Because the core computation of neural networks (various forms of summation of multiplications: convolution, matrix multiplication, einsum) are highly parallelizable numerical calculations, these chips are equipped with huge number of floating processing units (FPUs). Hence, the compute power of these specially designed hardware grew dramatically. It is reported that GPU price per flops dropped a factor of ten in just the last 4 years~\cite{gpu2019price} and flops per watts increased by 2 magnitude over the past 12 years~\cite{sun2019summarizing}. The widely available low-cost computation power is a major enabler for the success of neural networks. 

\textbf{Software} Software systems supporting neural networks evolved together with the advancement of the underlying hardware~\cite{dean2012large,bastien2012theano,abadi2016tensorflow,paszke2017automatic}. While the accelerators are highly parallel compute machines, they are significantly more difficult to program directly. The frameworks made building neural networks easier and abstracted away many hardware specific details from the practitioners. They in turn rely on lower-level libraries to drive special hardware (accelerators) efficiently. E.g., CUDA~\cite{nickolls2008scalable} for Nvidia's GPUs, or XLA for Google's TPUs~\cite{xla}. These lower-level libraries are critical for achieving high efficiency using these special hardware.

\textbf{Parallelism in model training and inference} Modern neural networks make extensive use of a cluster of machines for training and inference, each of which equiped with several accelerators.  Data parallelism~
\cite{krizhevsky2012imagenet} is the most commonly used approach and is supported by major frameworks (TensorFlow~\cite{abadi2016tensorflow}, PyTorch~\cite{pytorch2017}, JAX~\cite{jax2018github,frostig2018mlsys}), where devices run the same program with different input data and combine their local gradients before the weight updates. Model parallelism on the other hand, partitions computation beyond the input batch, which is needed to build very large models. For example, pipelining~\cite{gpipe19,harlap2018pipedream} splits a large model's layers into multiple stages, while operator-level partitioning~\cite{shazeer2018mesh,jia2018beyond} splits individual operators into smaller parallel operators. \codename used a type of operator-level partitioning to scale our model to a large number of parallel experts.

\textbf{Automated parallelism} Because programming in a distributed heterogeneous environment is challenging, particularly for high-level practitioners, deep-learning frameworks attempt to alleviate the burden of their users from specifying how the distributed computation is done. For example, TensorFlow~\cite{abadi2016tensorflow} has support for data parallelism, and basic model parallelism with graph partitioning by per-node device assignment. Mesh TensorFlow~\cite{shazeer2018mesh} helps the user to build large models with SPMD-style per-operator partitioning, by rewriting the computation in a Python library on top of TensorFlow; in comparison, our approach partitions the graph in the compiler based on light-weight annotations without requiring the user to rewrite the model. FlexFlow~\cite{jia2018beyond} uses automated search to discover the optimal partition of operators in a graph for better performance; while it focuses on determining the partitioning policy, our SPMD partitioner focuses on the mechanisms to transform an annotated graph. Weight-update sharding~\cite{xu2020automatic} is another automatic parallelization transformation based on XLA, which mostly focuses on performance optimizations for TPU clusters, and conceptually can be viewed as a special case for \codename. Zero~\cite{rajbhandari2019zero} presents a set of optimizations to reduce memory redundancy in parallel training devices, by partitioning weights, activations, and optimizer state separately, and it is able to scale models to 170 billion parameters; in comparison, \codename is more general in the sense that it does not distinguish these tensors, and all of those specific partitioning techniques can be supported by simply annotating the corresponding tensors, allowing us to scale to over 1 trillion parameters and explore more design choices.

\textbf{Conditional Computation and Machine Translation} 
Conditional computation~\cite{bengio2015conditional,shazeer2017outrageously,Elbayad2020DepthAdaptiveT,bapna2020controlling} premises that the examples should be routed within the network by activating an input dependent sub-network. The routing depends (or conditions) on certain criterion and without the loss of generality, can be any of the following: estimated difficulty of the example \cite{lugosch2020surprisaltriggered}, available computation budget \cite{Elbayad2020DepthAdaptiveT,bapna2020controlling}, or more generally a learned criterion with sparsity induced mixture of experts \cite{shazeer2017outrageously}. We extend sparsely gated mixture of experts \cite{shazeer2017outrageously} due to its flexibility and ease of scaling to state of the art neural sequence models, Transformers \cite{vaswani2017attention}, to satisfy training efficiency.

\section{Conclusion}
% Whatever alternative version of abstract 
In this paper, we introduced \codename, a deep learning module that partitions computation at scale automatically. GShard operates with  lightweight sharding annotations required in the user model code only and delivers an easy to use and flexible API for scaling giant neural networks.  We applied \codename to scale up Transformer architecture with Sparsely-Gated Mixture-of-Experts layers (MoE Transformer) and demonstrated a 600B parameter multilingual neural machine translation model can efficiently be trained in 4 days achieving superior performance and quality compared to prior art when translating 100 languages to English with a single model. In addition to the far better translation quality, MoE Transformer models trained with GShard also excel at \textit{training efficiency}, with a training cost of 22 TPU v3 core years compared to 29 TPU years used for training all 100 bilingual Transformer baseline models. Empirical results presented in this paper confirmed that scaling models by utilizing conditional computation not only improve the quality of real-world machine learning applications but also remained practical and sample efficient during training. Our proposed method presents a favorable scalability/cost trade-off and alleviates the need for model-specific frameworks or tools for scaling giant neural networks. Together, our results help to elucidate a realistic and practical way forward for neural network scaling to achieve better model quality.

We have learned several lessons from our study. Our results suggest that progressive scaling of neural networks yields consistent quality gains, validating that the quality improvements have not yet plateaued as we scale up our models. While the results in this paper consolidate that model scaling is a must in deep learning practitioners' toolbox, we also urge practitioners to strive for training efficiency. To this end, we identified factors that affect the training efficiency and showed their implications on downstream task quality. We demonstrated how the neural networks built with conditional computation yield a favorable trade-off between scale and computational cost. In practice such critical design decisions allowed us to enjoy experimental cycles of not months or weeks, but only days to train models in the order of magnitude of trillion parameters. 

Further, having a proper abstraction layer that separates model description from parallelization implementation, allows model developer to focus on network implementation, leaving \codename to partition the computation graphs automatically and generate programs that run on all devices in parallel.
We found that generating a single program that is general enough to express computation on all underlying parallel devices is the key to compile scalably. The traditional way of generating multiple dedicated programs for different partitions results in explosive compilation time when scaling to thousands of partitions. To address this complexity, we introduced various compiler renovations based on SPMD sharding that allows any tensor dimension to be partitioned. As a takeaway, we emphasize that model scaling and training efficiency should go hand-in-hand; and algorithmic improvements such as conditional computation when coupled with easy to use interfaces can effectively utilize large computational power. 

Lastly, our experimental results empirically support that, mere parameter counting does not always correlate with the effective capacity of the models at scale \cite{li2018measuring,maddox2020rethinking}. Comparison of the models should also account in the nature of the problem, i.e. massively multi-task setting with a heavy training data imbalance across tasks as in our case, and control the factors affecting different operation modes of the networks, i.e. capacity bottleneck vs positive transfer.

\section*{Acknowledgements}

We would like to thank the Google Brain and Translate teams for their useful input and insightful discussions, entire XLA and Lingvo development teams for their foundational contributions to this
project. In particular
Youlong Cheng, % first impl
Naveen Arivazhagan, Ankur Bapna, % m4 contribs
Ruoming Pang, % review
Yonghui Wu, % review
Yuan Cao, % review
David Majnemer, % review
James Molloy, % sharding APIs
Peter Hawkins, Blake Hechtman, Mark Heffernan, Dimitris Vardoulakis, % discussion on SPMD ideas
Tamas Berghammer, Marco Cornero, % sharding propagation
Cong Liu, % XLA collectives
Tong Shen, Hongjun Choi, Jianwei Xie, % TF/XLA bridge weight sharding
Sneha Kudugunta, % sentence-level MoE
and Macduff Hughes. % TPUs

\bibliographystyle{unsrt}
\bibliography{paper}

\appendix
\section{Appendix}
\label{sec:appendix}

% \todo{lepikhin: move EGC notation here}
% See "Notation and MoE layer" in model.tex, I tried to make is concise, it's hard to describe MoE layer without it.

\subsection{Decoding with Flat Beam Search}

During decoding, we use beam search with length normalization similar to
\cite{wu2016google}. Decoding is auto-regressive and generates the target sequence one token at a
time, so for an output of length $m$ the decoder layer stack is executed
$m$ times, sequentially. In particular for each decoder MoE layer there
are dispatch/combine operations, which require cross-device communication. 
Inference utilizes same cluster with same number of devices as training.

% flat beam search
During beam search we flatten the beam hypotheses into a single sequence
which contains all underlying tokens interleaved,
and we modify decoder self-attention mask so that each hypothesis
only has attention to appropriate positions in the joint flat sequence.
We apply the same transformation to key/value tensors maintained
by each decoder self-attention layer.
This allows us to avoid reordering previously computed
attention key/values after each beam expansion. Instead,
we only reorder the $0/1$ mask representing the current active hypotheses.
However, attention becomes $k$ times longer.

This trade-off can be positive or negative depending on implementation details.
As explained in \cite{shazeer2019fast}, memory bandwidth limits
are important for incremental decoding with Transformer models.
From this point of view, by flattening the beam we replace two 
operations with low compute/memory ratio
(attention dot product and key/value reordering)
with a single operation with a slightly higher compute/memory ratio
(attention dot product over a longer sequence with more keys),
but with the same total amount of memory it has to access.

% Maybe put fast distillation into appendix
% \subsubsection*{Fast distillation}

\subsection{Machine Translation Experiments Details}
\label{sec:mt-more}

In our Machine Translation experiments MoE Transformer models shared \textit{a)} 1024 Transformer model dimension 
\textit{b)} 8192 Feed Forward and MoE hidden dimension;
\textit{c)} 16 heads in multi-head attention;
\textit{d)} 128 attention key and value dimension; and
\textit{e)} 0.1 input, residual and attention dropout rate.
% \begin{enumerate*}[label=\textit{\alph*}), 
%       itemjoin={{; }},itemjoin*={{; and }}]
% \item 1024 Transformer model dimension
% \item 8192 Feed Forward and MoE hidden dimension
% \item 16 heads in multi-head attention
% \item 128 attention key and value dimension
% \item 0.1 input, residual and attention dropout rate
% \end{enumerate*}. 

We used the Adafactor~\cite{Shazeer2018AdafactorAL} optimizer with
\textit{a)} factored second-moment estimation;
\textit{b)} first moment decay $\beta_1=0.0$;
\textit{c)} second moment decay $\beta_2=0.99$ with $1 - t^{-0.8}$ schedule;
\textit{d)} update clipping threshold of 1.0; and
\textit{e)} 1.0 learning rate with square root decay after 10k training steps.
% \begin{enumerate*}[label=\textit{\alph*}), 
%       itemjoin={{; }},itemjoin*={{; and }}]
% \item factored second-moment estimation
% \item first moment decay $\beta_1=0.0$
% \item second moment decay $\beta_2=0.99$ with $1 - t^{-0.8}$ schedule
% \item update clipping threshold of 1.0
% \item 1.0 learning rate with square root decay after 10k training steps
% \end{enumerate*}. 

We used SentencePiece~\cite{Kudo2018SentencePieceAS} subword tokenizer with a single multilingual vocabulary for source-side spanning 102 languages of size 64000, and English-only target-side vocabulary of size 32000.

\subsection{General Sharding API}\label{sec:appendix-shard-api}
In addition to the two common APIs (\texttt{replicate()} and \texttt{split()}) for sharding listed in Section~\ref{sec:spmd_annotation}, users or the compiler may use a more advanced sharding strategy to minimize data transfers.
    
\textbf{shard(tensor, device\_assignment)} annotates \texttt{tensor} to be partitioned with the provided device assignment, and returns the annotated tensor. We use \textit{device assignment}, a multi-dimensional integer array, to represent how the split is done. \texttt{device\_assignment} has the same rank as the data tensor; its element count is the total number of partitions, and each element is the ID of the device that occupies the corresponding data slice. For example, a 3D tensor with shape $[3,16,64]$ with device assignment shape $[1,2,4]$ will have partition shape $[3,8,16]$, and the order of elements in \textit{device assignment} determines which slice each partition occupies.

\begin{figure}[hb]
\begin{center}
\includegraphics[width=0.98\textwidth]{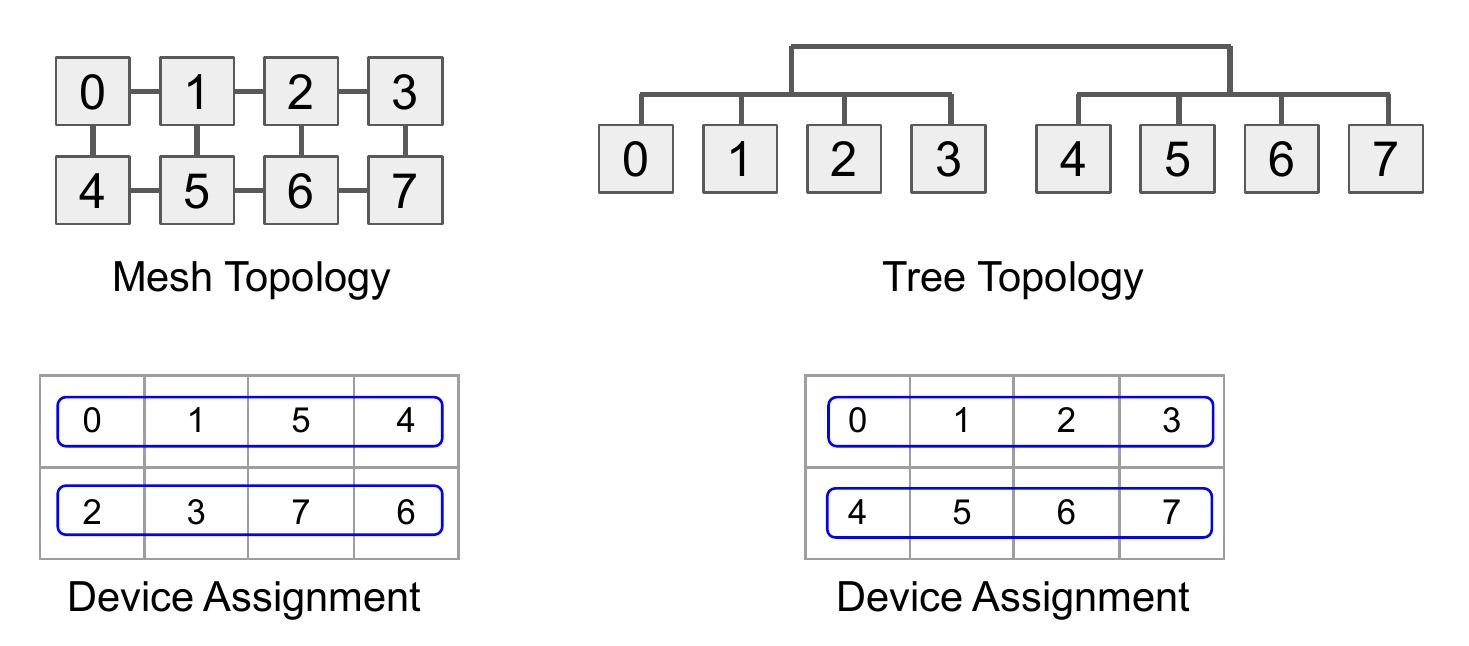}
\caption{An example of two different \textit{device assignments} based on the device topology. A 2D tensor is split by 2x4 partitions and the communication pattern is between partitions along the rows of the tensor. The numbers represent device ids.}
\label{fig:device_assignment}
\end{center}
\end{figure}

Since data movement across devices critically affects the parallel execution performance, it is important to consider the target device topology as well as the communication between partitions of the tensor when assigning device ids in the \textit{device assignment} for maximum performance. Figure~\ref{fig:device_assignment} shows two different \textit{device assignments} based on the device topology and the row-wise communication pattern on the tensor.

\subsection{SPMD Partitioning for Convolution and Window-Based Operators}\label{sec:appendix-spmd-conv}
\codename is able to partition spatial dimensions in convolutions, and general enough to support use cases like giant images~\cite{spatial-partitioning}. To spatially shard a convolutional layer, we can use the sharding API in the following way.

\begin{lstlisting}[escapeinside={&}{&},escapechar=\%]
  # Partition input images [N,C,H,W] along W spatial dimension
%\Hilight% inputs = split(inputs, 3, D)
  # Replicate the kernel
%\Hilight% kernel = replicate(kernel)
  conv = conv2d(inputs, kernel)
  ...
\end{lstlisting}

\codename will then propagate the sharding on the spatial dimension to other layers and the backward pass. The rest of section discusses the specific complexity to partition \texttt{Convolution} and similar operators.
There are several window-based operations (e.g., \texttt{Convolution}, \texttt{ReduceWindow}), and they all require some type of halo exchange since data may be shared between windows. We use the \texttt{CollectivePermute} operator to exchange halo data between partitions, but one complication is that the halo size may be different across partitions whereas \texttt{CollectivePermute} needs to be statically shaped.

% \begin{figure}
% \begin{center}
% \includegraphics[width=0.98\textwidth]{spmd_convolution.png}
% \caption{SPMD partitioning of a convolution with halo exchange}
% \label{fig:spmd_convolution}
% \end{center}
% \end{figure}

% Figure~\ref{fig:spmd_convolution} shows an example of a \texttt{Convolution} where the input and the output are partitioned across 2 devices and the kernel is replicated. The kernel size is 3 and the left/right padding is 1 (i.e., \texttt{SAME} padding), so each partition needs to read 1 column from the input of the other partition for the halo data. The SPMD transformation of this example goes through following steps:

% \begin{itemize}
%     \item To extract the halo data, use \texttt{DynamicSlice} to select the right-most column (from Partition 0) or the left-most column (from Partition 1) of the partitioned input.
%     \item Send the result of the \texttt{DynamicSlice} to the other partition using \texttt{CollectivePermute} with source/target pairs (0:1) and (1:0).
%     \item Partition 0 needs zero padding on the left and the halo data on the right. On the other hand, Partition 1 needs zero padding on the right and the halo data on the left. Therefore, call \texttt{Select} twice with a zero-valued tensor and the result from \texttt{CollectivePermute} to choose which one to put left and right for each partition.
%     \item \texttt{Concatenate} the result of the two \texttt{Select}s and the original partitioned input, and call \texttt{Convolution} with both low and high padding set to 0. 
% \end{itemize}

We first introduce the window configurations that the SPMD partitioner has to consider. Each spatial dimension in the convolution has the following set of configurations.
\begin{itemize}
    \item \textbf{Stride} is the distance (in number of elements) that the window moves to produce the next output element.
    \item \textbf{Low/high padding} is the number of elements padded to the low/high end of the dimension in LHS (base).
    \item \textbf{Base dilation} is the dilation factor of the LHS, i.e., one plus the number of elements padded between every element (excluding low/high padding). No base dilation means the value is set to 1.
    \item \textbf{Window dilation} is one plus the number of elements padded between every element in the RHS (window).
\end{itemize}

\begin{figure}
\begin{center}
\includegraphics[width=0.6\textwidth]{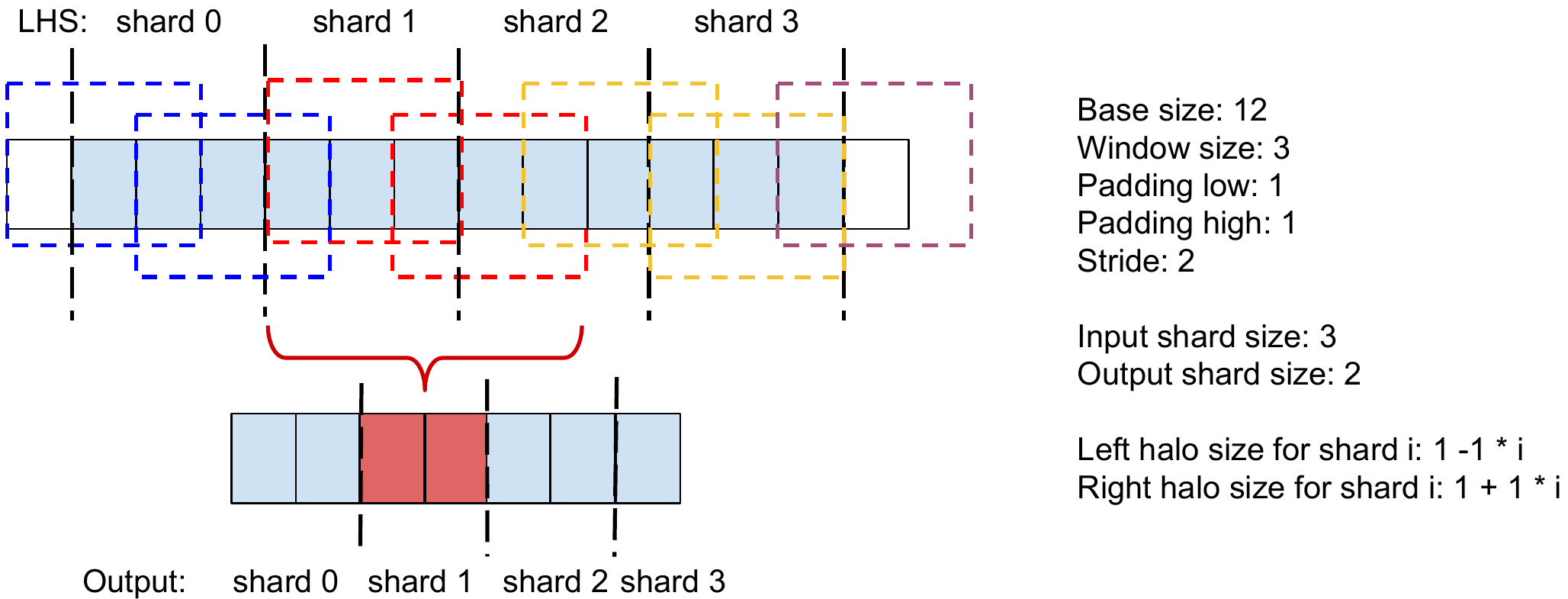}
\caption{Convolution with non-constant halo size.}
\label{fig:conv_nonconstant_halo}
\end{center}
\end{figure}

\paragraph{Non-constant halo size.} We demonstrate that non-constant halo size is common using a simple example, which does not have dilation. Figure~\ref{fig:conv_nonconstant_halo} shows a 4-way partitioned convolution, where the right halo sizes for the partitions are (1, 2, 3, 4) and can be expressed as a linear function of the partition ID: $partition\_id + 1$. Partition 1 is in charge of generating 2 output elements (red cells), which means that the partition needs to get 0 elements from Partition 0, and 2 elements from Partition 2 (area covered by two dotted red windows).

\begin{figure}
\begin{center}
\includegraphics[width=0.8\textwidth]{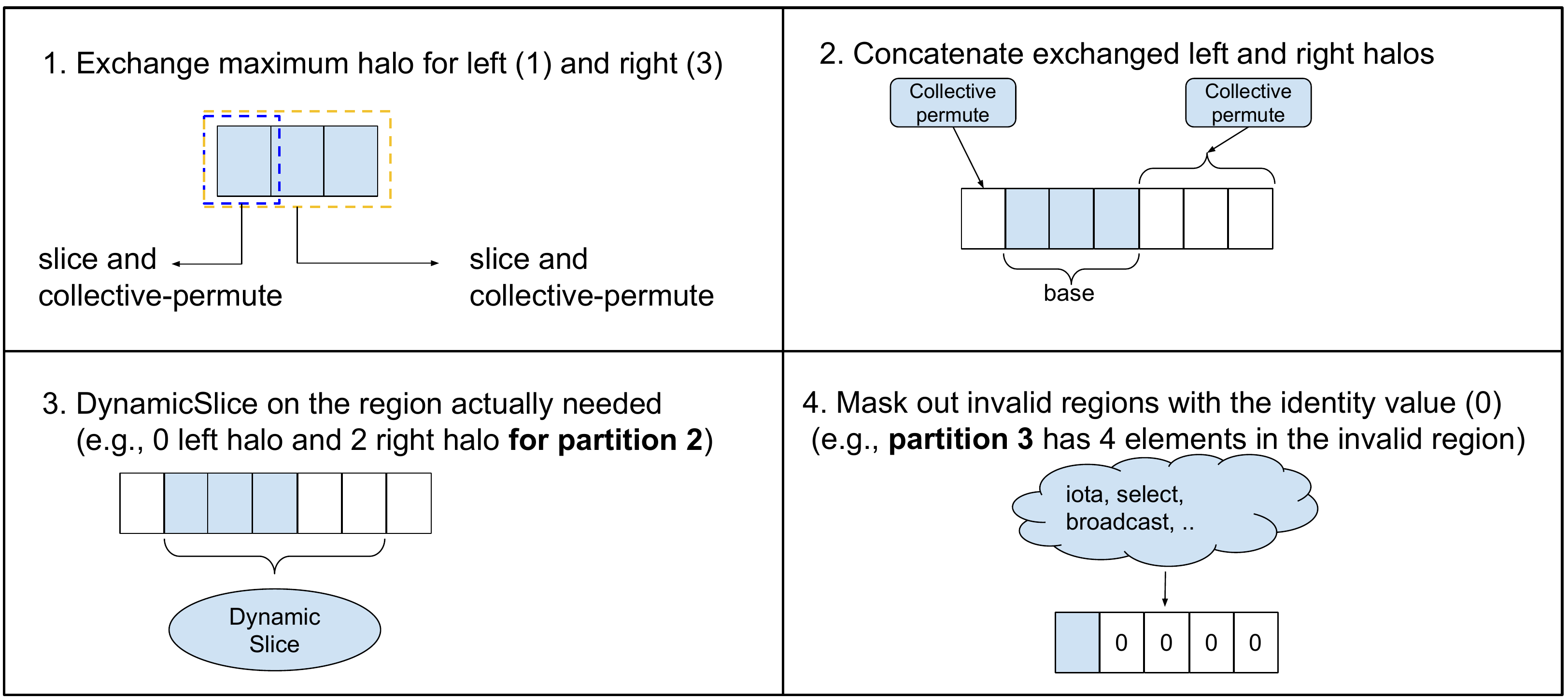}
\caption{Sequence of operations for a general halo exchange.}
\label{fig:conv_handling}
\end{center}
\end{figure}

Figure~\ref{fig:conv_handling} describes the sequence of operations for a general halo exchange. First, we calculate the maximum size of left and right halo across partitions and perform the halo exchange of the maximum size (Steps 1 and 2). Since some partitions may have excessive halos than needed, we use \texttt{DynamicSlice} (based on the partition ID) to slice off the valid region for the current partition (Step 3). Finally, some partitions may include garbage values (e.g., halos from out-of-range input data), so we apply masking as described in Section~\ref{sec:spmd-general-cases}.

\begin{figure}
\begin{center}
\includegraphics[width=\textwidth]{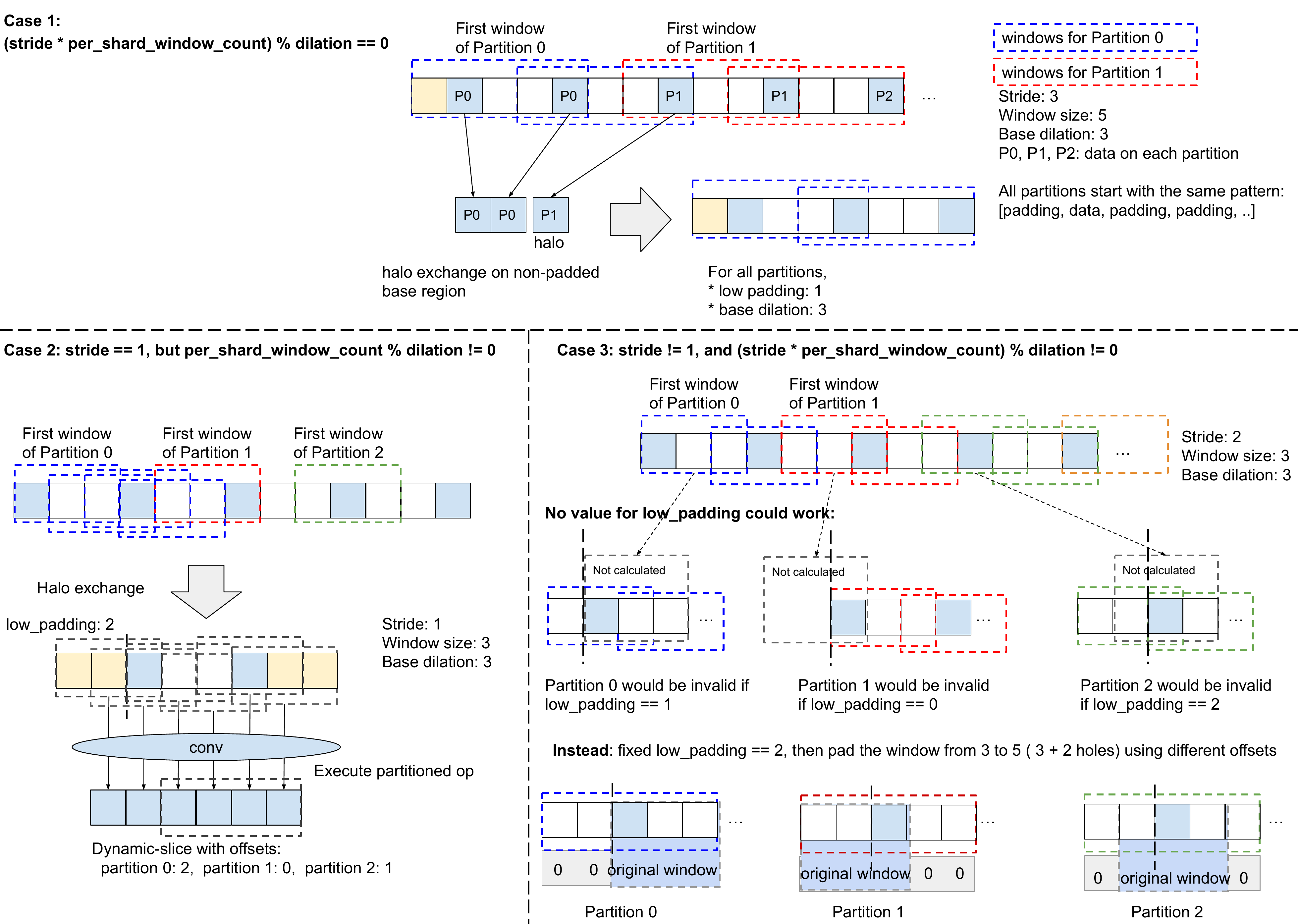}
\caption{Partitioned convolution with base dilation.}
\label{fig:base_dilation}
\end{center}
\end{figure}

\paragraph{Base dilation.} Base dilation adds additional complexities to halo exchange, since the offset of each partition may be positioned at the dilation holes, and also low/high padding is applied after dilation, which makes the edges have different behavior than the interior elements. We handle base dilation in 3 cases (Figure~\ref{fig:base_dilation}).

\begin{itemize}
    \item $stride \times per\_shard\_window\_count$ is divisible by $dilation$, where $per\_shard\_window\_count$ is the number of windows to be processed by each partition (i.e., the number of output elements for each partition). This condition guarantees that all partitions start with the same number of (interior or low) padding elements before the first data element in the LHS, so that we can use the same low padding. Halo exchange occurs on the non-dilated/non-padded base region, and the limit index of required data for Partition $i$ can be calculated as below.
    $$
         \frac{stride \times per\_shard\_window\_count \times i + window\_size - low\_pad + dilation - 1}{dilation},
    $$
    which determines the right halo size. Because $stride \times per\_shard\_window\_count$ is divisible by $dilation$, it can be simplified as $a \times i + b$, where $a$ and $b$ are both constants.

    \item $stride == 1$ but $per\_shard\_window\_count$ is not divisible by $dilation$. In this case, the low padding on different partitions are different, but it is a static configuration in windowed operations, which can’t be specialized for each partition for SPMD execution. Using \texttt{Pad} and \texttt{DynamicSlice} on the operand also would not work, because those operators would be applied before dilation, so everything would be multiplied by the dilation factor. Fortunately, with $stride == 1$, all positions on the padded and dilated base region are valid window starts, and we can use the maximum low padding on all partitions to ensure that each partition calculates all required windows, then do a \texttt{DynamicSlice} on the output of the partitioned windowed operator to remove unnecessary data. The limit index of required data on the non-padded base region for Partition $i$ is same as before,
    $$
      \frac{per\_shard\_window\_count \times i + window\_size - low\_pad + dilation - 1}{dilation},
    $$
    but cannot be simplified to $a \times i + b$.

    \item $stride\neq 1$ and $stride \times per\_shard\_window\_count$ is not divisible by $dilation$. If neither of the above conditions are true, different partitions could start with different number of padding elements, and not all offsets are valid window starts. Consider the last example in Figure~\ref{fig:base_dilation}. Whatever low padding we chose, some partition will be invalid, because the valid windows could be skipped since $stride\neq  1$. A solution to this problem is to pad the window in addition to padding the base area. We can use the maximum low padding required by the partitions on the base area, then increase the window size by that low padding amount. However, the low and high padding amounts on the window vary on different partitions, which can be implemented by a \texttt{Pad} followed by a \texttt{DynamicSlice}. The window padding is used to mask off the unaligned elements in the base area, so that the start of the non-padding window element will be aligned with the desired start in the base area for each partition.
\end{itemize}

\paragraph{Window dilation.} If the RHS is replicated, window dilation only affects the effective window size when partitioning the operator based on its LHS. If the dilated RHS is also partitioned, which typically occurs in the gradient computation of strided convolutions, handling window dilation is still simpler than handling base dilation, because there is no low/high padding on the RHS. We skip the details of the implementation.

\end{document}